\documentclass[11pt]{article}

\usepackage[preprint]{acl}

\usepackage{times}
\usepackage{latexsym}

\usepackage[T1]{fontenc}

\usepackage[utf8]{inputenc}

\usepackage{microtype}

\usepackage{inconsolata}
\usepackage{microtype}
\microtypecontext{spacing=nonfrench}
\usepackage{amssymb}
\usepackage{hyperref}
\usepackage{longtable}
\usepackage{arydshln}
\usepackage{bm}      
\usepackage{booktabs}
\usepackage{array}
\usepackage{makecell}
\usepackage{amsmath, amssymb, amsthm}
\usepackage{cleveref}  
\usepackage{amsmath, etoolbox}
\usepackage{algorithm}
\usepackage{algorithmic}
\usepackage{wrapfig}
\usepackage{makecell}
\usepackage{graphicx}
\usepackage{ragged2e} 
\usepackage{graphicx}
\usepackage{xspace}
\usepackage{multirow}
\usepackage{pifont} 
\usepackage{colortbl} 
\PassOptionsToPackage{table}{xcolor} 
\usepackage{xcolor}
\usepackage{subcaption}
\usepackage{comment}
\usepackage{tabularx}
\usepackage{microtype}      
\usepackage{xcolor}         
\usepackage{hyperref}
\usepackage{longtable}
\usepackage{arydshln}
\usepackage{bm}      
\usepackage{booktabs}
\usepackage{array}
\usepackage{makecell}
\usepackage{amsmath, amssymb, amsthm}
\usepackage{cleveref}  
\usepackage{amsmath, etoolbox}
\usepackage{subcaption}
\usepackage{ragged2e} 
\usepackage{graphicx}
\usepackage{bbm}
\usepackage{xspace}
\usepackage{multirow}
\usepackage{pifont} 
\usepackage{colortbl} 
\usepackage[most]{tcolorbox}
\newcommand{\methodname}{\textsc{Mvpo}}
\newcommand\secref[1]{\S\ref{#1}}

\title{Learning from Viable Failure Prefixes: \underline{M}ilestone \underline{V}iability Potential \underline{P}olicy \underline{O}ptimization for Long-Horizon LLM Agents}

\author{Qi Zhou\textsuperscript{$\dagger$}, Yuanfan Li\textsuperscript{$\dagger,\ast$} \\
  Xi'an Jiaotong University \\
  \textsuperscript{$\dagger$} Equal contribution, \textsuperscript{$\ast$} Correspondence to: \texttt{liyuan7716@gmail.com}
}

\begin{document}
\maketitle
\begin{abstract}
Long-horizon LLM agents require reinforcement learning methods that can assign credit to intermediate decisions under sparse and delayed rewards. Existing group-based methods such as GRPO and GiGPO alleviate this issue by comparing rollout returns or repeated anchor states, but they still fail when the compared returns have no variation. We identify this failure mode as \emph{zero-credit failure}: during early training, many failed rollouts contain useful prefixes, yet existing methods assign them no task-discriminative advantage. To address this issue, we propose \textbf{\underline{M}ilestone \underline{V}iability Potential \underline{P}olicy \underline{O}ptimization} (\methodname{}), a potential-routed policy optimization algorithm that learns from viable failure prefixes. \methodname{} estimates prefix potential over Union-Find viability regions, repairs zero-credit groups with potential-difference advantages, and attenuates the potential branch according to relative performance progress. Experiments with Qwen2.5-1.5B-Instruct show that \methodname{} outperforms eight strong baselines, including GRPO and GiGPO. Under the same training length, \methodname{} improves over the GiGPO baseline by \textbf{+4.4} success points on ALFWorld and \textbf{+5.3} on WebShop, while adding only \textbf{0.16\%--0.20\%} advantage-construction overhead.
\end{abstract}

\section{Introduction}

\begin{figure}[t]
    \centering
    \resizebox{0.48\textwidth}{!}{\includegraphics{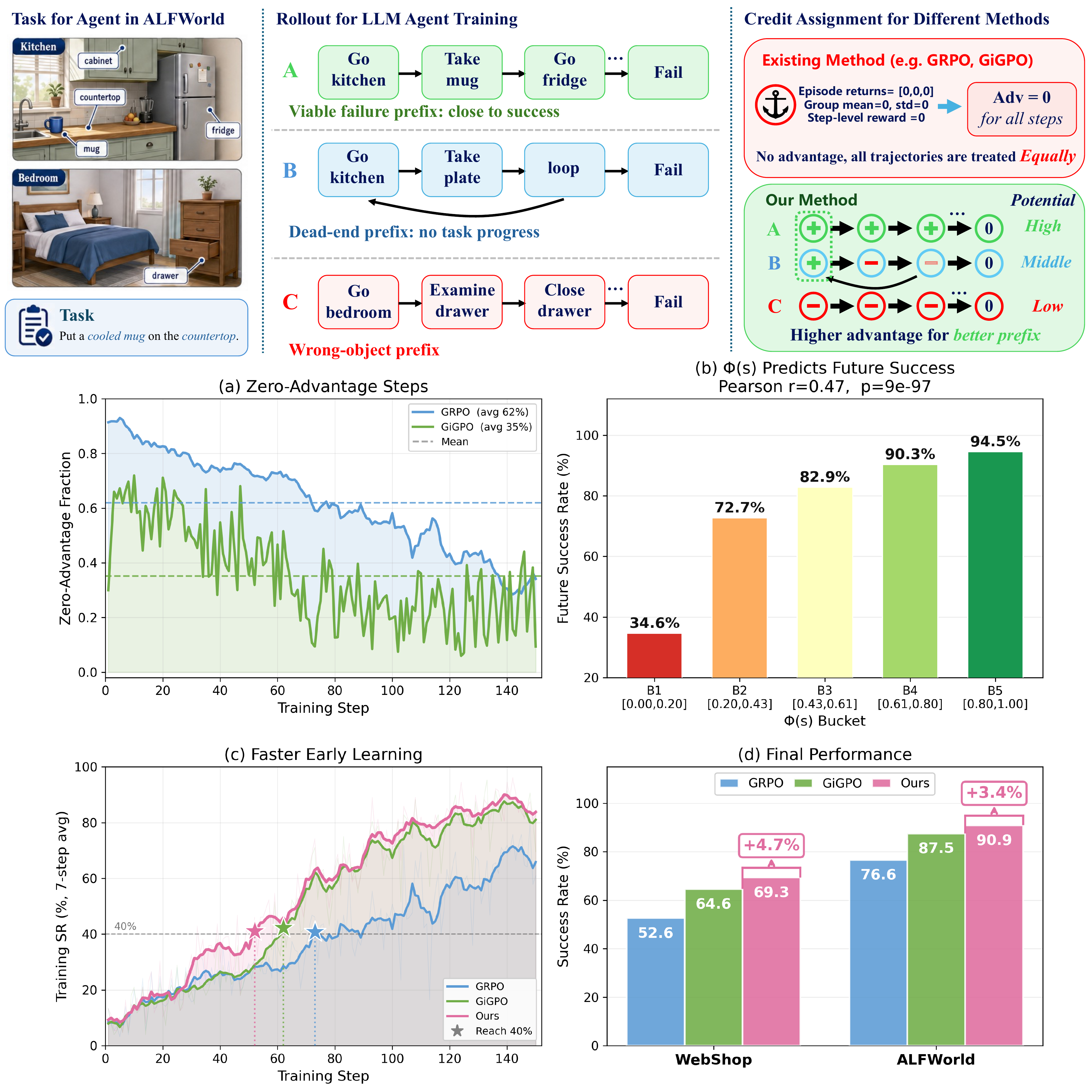}}
    \caption{
    \textbf{Motivation and overview of \methodname.}
    \textbf{Top}: an ALFWorld case of \emph{zero-credit failure}: all rollouts fail, making GRPO/GiGPO assign no task-discriminative advantage, while \methodname{} identifies viable prefixes using prefix potential.
    \textbf{Bottom (a)}: zero-advantage steps affect over 90\% of GRPO steps and about 65\% of GiGPO steps in early WebShop training.
    \textbf{Bottom (b)}: higher prefix potential predicts higher future success on ALFWorld.
    \textbf{Bottom (c)}: \methodname{} reaches 40\% success earlier during training.
    \textbf{Bottom (d)}: \methodname{} improves final success over GiGPO on both benchmarks.
    }
    \label{fig:compare}
    \vspace{-1.5em}
\end{figure}

Large language models (LLMs) have achieved remarkable progress across language understanding, reasoning, and generation tasks~\cite{vaswani2017attention,brown2020language,singh2025openai}, and are increasingly being developed as autonomous agents that interact with external environments, execute multi-step plans, and solve long-horizon decision-making problems~\cite{bai2022training,shao2024deepseekmath,feng2025group}. 
Such \emph{agentic} LLMs have been studied in embodied environments, web navigation, tool-use tasks, and application-centered workflows, where success requires a sequence of environment-dependent actions rather than a single response~\cite{shridhar2021alfworld,yao2022webshop,trivedi2024appworld,wei2025webagent}. 
Training these agents with reinforcement learning (RL) is challenging because rewards are often sparse and delayed: most early rollouts fail before reaching the goal, and non-zero feedback is observed only after completing the task. 
Recent critic-free policy optimization algorithms attempt to mitigate this issue by comparing multiple rollouts under the same prompt. 
GRPO~\cite{shao2024deepseekmath} computes group-relative advantages from trajectory-level returns, but its credit signal remains tied to final episode outcomes. 
GiGPO~\cite{feng2025group} further adapts group-based optimization to agentic tasks by constructing step-level advantages at repeated \emph{anchor states}, comparing subsequent returns of rollouts that visit the same state. 
Although this provides denser supervision than trajectory-level GRPO, we find that it still fails to eliminate sparse-reward failure in long-horizon agentic training.

The key issue is that both rollout-level and anchor-level comparisons require return variation, which is especially scarce at the beginning of training.
When most sampled rollouts fail, GRPO receives nearly identical trajectory returns and therefore assigns no task-discriminative relative advantage.
GiGPO alleviates this issue by comparing continuations from repeated anchor states, but its step-level advantage still collapses whenever all continuations from an anchor have identical returns.
We call these cases \emph{zero-credit groups}: groups in which existing credit estimators cannot distinguish useful decisions from useless ones because the observed returns contain no variation.
As shown in Figure~\ref{fig:compare}(a), zero-credit groups dominate the early training stage: over 90\% of GRPO steps and more than 60\% of GiGPO steps have zero advantage.
This means that a large fraction of early interactions provides no task-discriminative policy-gradient signal, slowing down the initial rise of the training curve.
Thus, early agent training is sparse not only at the trajectory level, but also at the prefix level.

To understand the cause of this prefix-level sparsity, we examine what is hidden inside failed rollouts.
Figure~\ref{fig:compare} (top) shows an ALFWorld task, ``put a cooled mug on the countertop,'' where three rollouts all fail with zero returns.
Yet their prefixes differ substantially: one rollout reaches the kitchen, takes the mug, and approaches the fridge, while the others pick up a wrong object or explore irrelevant locations.
GRPO and GiGPO treat these rollouts as equally uninformative once their returns are all zero.
This motivates our view that failed trajectories should not be uniformly discarded; their viable prefixes should be identified and reinforced.

Motivated by this diagnosis, we propose \textbf{\underline{M}ilestone \underline{V}iability Potential \underline{P}olicy \underline{O}ptimization} (\methodname), a potential-routed LLM agent training algorithm that learns from viable failure prefixes.
Instead of assigning credit only from final trajectory outcomes, \methodname{} estimates a \emph{prefix potential}, which measures whether an intermediate state remains on a promising path toward task completion.
For example, finding the apple or reaching the fridge should have higher potential than entering a shelf loop or holding an irrelevant object.
To estimate this signal robustly, \methodname{} uses a Union-Find structure to merge semantically equivalent prefixes across rollouts into \emph{viability regions}, where each region aggregates future success, loop frequency, and adaptive milestone progress.
During optimization, \methodname{} routes the step-level signal according to credit availability: it preserves comparative anchor advantages when they are informative, and repairs zero-credit groups with potential-difference advantages that reward transitions toward higher-potential regions and suppress regressions or dead ends.
Finally, \methodname{} learns milestone weights from rollout statistics and attenuates the potential branch according to relative performance progress, so that potential-based credit dominates early sparse-reward training but gradually gives way to anchor-based credit later.

Figure~\ref{fig:compare}(b--d) validates both the mechanism and effectiveness of \methodname{}.
On ALFWorld, future success increases from 34.6\% in the lowest-potential bucket to 94.5\% in the highest-potential bucket, showing that prefix potential captures meaningful task progress.
Using this signal to repair zero-credit steps, \methodname{} reaches 40\% ALFWorld success at step 52, earlier than GiGPO at step 62 and GRPO at step 73.
Across ALFWorld~\cite{shridhar2021alfworld} and WebShop~\cite{yao2022webshop}, \methodname{} outperforms eight strong baselines, including prompt-based agents, actor-critic RL, GRPO~\cite{shao2024deepseekmath}, and GiGPO~\cite{feng2025group}.
Under the same training length, it improves ALFWorld/WebShop success over GiGPO by \textbf{+4.4}/\textbf{+5.3} points with Qwen2.5-1.5B-Instruct and \textbf{+1.1}/\textbf{+2.7} points with Qwen2.5-7B-Instruct, while adding only \textbf{0.16\%--0.20\%} advantage-construction overhead.
Our contributions are as follows:

\begin{itemize}
    \item \textbf{Zero-Credit Prefix Failure Diagnosis.}
    We identify a key limitation of existing group-based policy optimization methods in long-horizon agent training: when rollout or anchor groups contain identical returns, their advantages collapse to zero. This prevents GRPO and GiGPO from distinguishing viable failure prefixes from completely unproductive trajectories during early sparse-reward training.

    \item \textbf{Potential-Routed Prefix Credit Repair.}
    We propose \methodname{}, a new potential-routed policy optimization algorithm that learns from viable failure prefixes. \methodname{} estimates prefix potential via Union-Find viability regions and repairs zero-credit groups with potential-difference advantages, while preserving anchor-based comparative credit when it is available.

    \item \textbf{Strong Long-Horizon Agent Performance.}
    We evaluate \methodname{} on ALFWorld and WebShop, where it outperforms eight strong baselines, including prompt-based agents, actor-critic RL methods, GRPO, and GiGPO. Under the same training length, \methodname{} improves ALFWorld/WebShop success over GiGPO by \textbf{+4.4}/\textbf{+5.3} points with Qwen2.5-1.5B-Instruct and \textbf{+1.1}/\textbf{+2.7} points with Qwen2.5-7B-Instruct, with only \textbf{0.16\%--0.20\%} additional advantage-construction overhead.
\end{itemize}

\section{Related Work}

\noindent \textbf{LLMs as autonomous agents.}
LLMs are increasingly used as autonomous agents that interact with environments, invoke tools, and solve multi-step tasks.
Early systems mainly relied on frozen LLMs with prompting, reasoning-action interleaving, memory, reflection, retrieval, and tool use, such as ReAct~\cite{yao2023react}, Reflexion~\cite{shinn2023reflexion}, WebGPT~\cite{nakano2021webgpt}, and Toolformer~\cite{schick2023toolformer}.
Agent benchmarks have expanded from embodied and text-based tasks to web navigation, shopping, mobile-device control, and application-centered workflows~\cite{shridhar2021alfworld,yao2022webshop,deng2023mind2web,zhou2024webarena,trivedi2024appworld,wen2024autodroid}.
These environments require long sequences of environment-dependent actions under delayed rewards, making policy learning harder than single-turn reasoning.
Prompting-based agents exploit pretrained knowledge, but provide limited mechanisms for improving policies from failed interactions, motivating supervised and reinforcement learning for LLM agents.

\noindent \textbf{Policy optimization for LLMs and agents.}
RL has been widely used to align and improve LLMs, from RLHF with PPO~\cite{ziegler2019fine,stiennon2020learning,ouyang2022training,schulman2017proximal} to critic-free estimators such as REINFORCE, RLOO, and GRPO~\cite{williams1992simple,ahmadian2024back,shao2024deepseekmath}.
Group-based algorithms such as Dr.GRPO, DAPO, CPPO, and GSPO~\cite{liu2025understanding,yu2025dapo,lin2026cppo,zheng2025group} further improve scalable LLM RL and have been applied to reasoning, search, and tool use~\cite{dong2025toolstar,dong2025arpo}.
For multi-turn agents, ArCHer~\cite{zhou2024archer} uses hierarchical RL, Agent Q~\cite{putta2024agent} combines MCTS with iterative fine-tuning, RAGEN/StarPO~\cite{wang2025ragen} studies trajectory-level agent optimization, Tree-GRPO~\cite{ji2025tree} derives process signals from tree-structured rollouts, GiGPO~\cite{feng2025group} constructs step-level advantages from repeated anchor states, SIRI~\cite{he2026siri} internalizes validated skills mined from successful rollouts, and ECPO~\cite{li2026ecpo} calibrates step-level credit through action-level shrinkage and variance-gated weighting.
These methods improve trajectory collection, internalize reusable skills, or densify credit from sparse outcomes, but group-relative estimators still rely on return variation within rollout or anchor groups.
When all rollouts fail or all anchor continuations have identical returns, their advantages collapse to zero, leaving viable prefixes inside failed trajectories unlearned.
In contrast, \methodname{} targets this zero-credit regime by routing such steps to prefix-potential advantages estimated over Union-Find viability regions, enabling agents to learn from viable failure prefixes while preserving comparative credit when available.
\section{Preliminaries}
\label{sec:preliminaries}

\noindent \textbf{Problem setup.}
We consider a long-horizon agentic RL setting where an LLM agent receives a task instruction $x\sim p(\mathcal{X})$ and interacts with an environment for multiple steps.
At step $t$, the agent observes state $s_t$, samples a textual action $a_t\sim\pi_\theta(\cdot\mid s_t,x)$, receives reward $r_t$, and transits to $s_{t+1}$.
An episode is a trajectory
\begin{equation}
    \tau=\{(s_1,a_1,r_1),\ldots,(s_T,a_T,r_T)\},
\end{equation}
with return $R(\tau)$.
In long-horizon agentic tasks, rewards are often sparse and delayed: most intermediate actions receive no direct supervision, and early rollouts frequently fail before reaching the goal.
The central challenge is therefore to assign useful credit to intermediate decisions, especially inside failed trajectories.

\noindent \textbf{Group-based credit estimation and zero-credit groups.}
Critic-free group-based RL estimates advantages by comparing multiple rollouts from the same task.
Given $N$ trajectories $\{\tau_i\}_{i=1}^{N}$ sampled from the old policy $\pi_{\theta_{\mathrm{old}}}$, GRPO~\cite{shao2024deepseekmath} computes a trajectory-level group-relative advantage:
\begin{equation}
\label{eq:grpo_adv}
    A^{\mathrm{epi}}_i
    =
    \frac{R_i-\mu_R}{\sigma_R+\epsilon},
    \quad
    \mu_R=\frac{1}{N}\sum_{j=1}^{N}R_j ,
\end{equation}
where $R_i=R(\tau_i)$ and $\sigma_R$ is the standard deviation of trajectory returns.
This avoids learning a value function, but all actions in the same trajectory share the same advantage.
When all rollouts fail and obtain identical returns, $\sigma_R=0$, so GRPO provides no task-discriminative signal.

GiGPO~\cite{feng2025group} improves this by constructing step-level credit at repeated \emph{anchor states}.
For trajectory $\tau_i$, the discounted future return from step $t$ is
\begin{equation}
\label{eq:future_return}
    G_{i,t}
    =
    \sum_{\ell=t}^{T_i}
    \gamma^{\ell-t}r_{i,\ell}.
\end{equation}
For an anchor state $s$, GiGPO collects its occurrence set
\begin{equation}
    \mathcal{I}_s=\{(i,t)\mid s_{i,t}=s\},
\end{equation}
and normalizes future returns within this anchor group:
\begin{equation}
\label{eq:anchor_adv}
    A^{\mathrm{anchor}}_{i,t}
    =
    \frac{G_{i,t}-\mu_s}{\sigma_s+\epsilon},
    \quad
    (i,t)\in\mathcal{I}_s .
\end{equation}
The final advantage is
\begin{equation}
\label{eq:gigpo_adv}
    \hat{A}^{\mathrm{GiGPO}}_{i,t}
    =
    A^{\mathrm{epi}}_i
    +
    \omega A^{\mathrm{anchor}}_{i,t}.
\end{equation}

Although GiGPO provides denser credit than GRPO, it still requires return variation inside anchor groups.
If all continuations from an anchor state have identical future returns, then $\sigma_s=0$ and the anchor advantage also collapses.
We call such rollout or anchor groups \emph{zero-credit groups}:
\begin{equation}
\label{eq:zero_credit_group}
    \mathrm{ZeroCredit}(g)
    \Longleftrightarrow
    \operatorname{Std}\big(\{G_u\mid u\in g\}\big)=0 .
\end{equation}
Zero-credit groups are especially common in early sparse-reward training, when most rollouts fail.
As a result, existing group-based PO methods treat viable failure prefixes and completely unproductive prefixes as equally uninformative.
This motivates \methodname{}, which repairs zero-credit groups with prefix-potential credit in Section~\ref{sec:method}.

\begin{figure*}[t]
    \centering
    \includegraphics[width=0.99\textwidth]{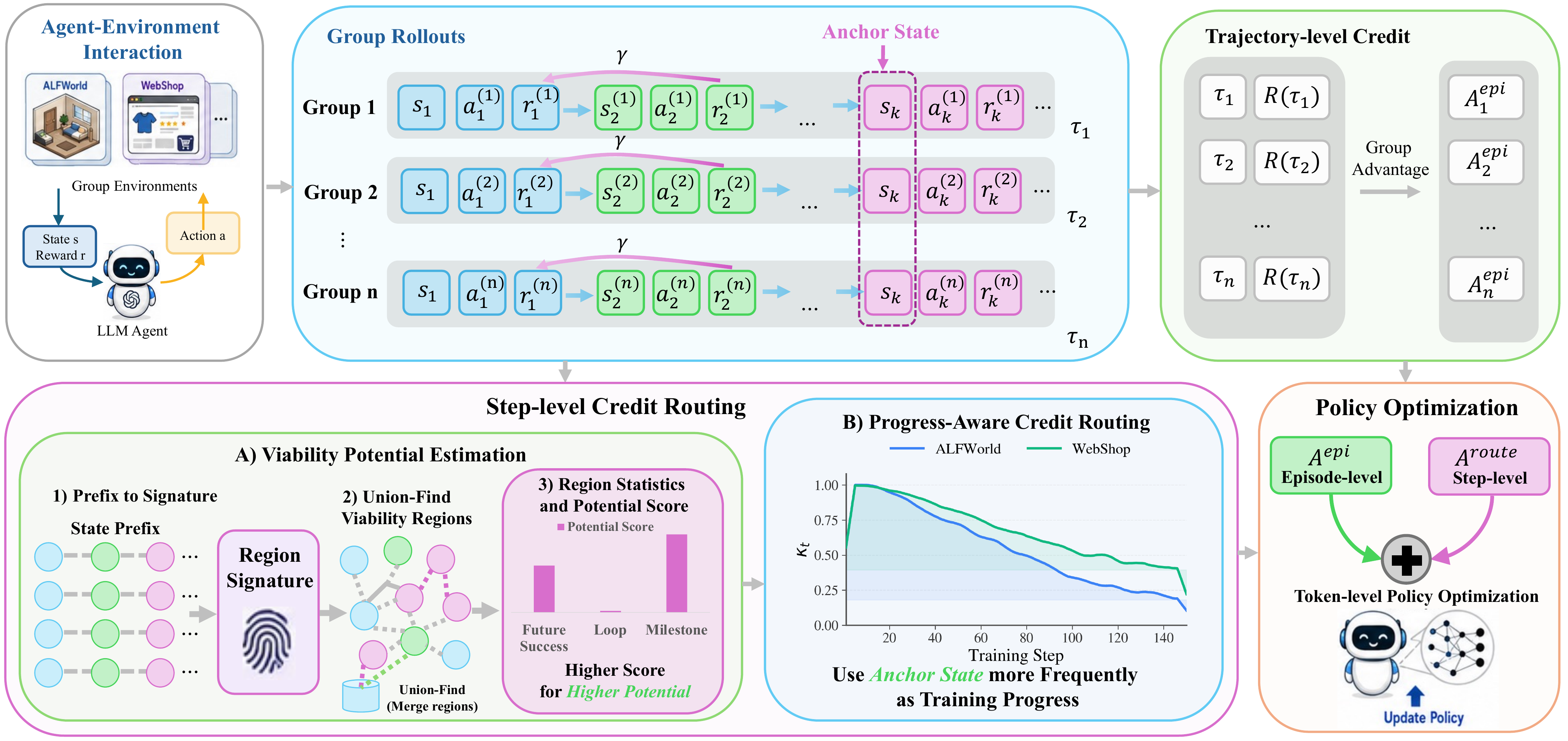}
    \caption{
    \textbf{Overview of \methodname{}.}
    \methodname{} first estimates prefix potential over Union-Find viability regions using future success, loop statistics, and adaptive milestone progress (\secref{sec:viability_potential}).
    It then routes step-level credit according to anchor-group availability: informative anchor groups use anchor-based comparative credit, while zero-credit groups are repaired with prefix-potential advantages (\secref{sec:credit_routing}).
    Finally, the routed step-level credit is combined with the episode-level group advantage for critic-free policy optimization (\secref{sec:method_optimization}).
    }
    \vspace{-1em}
    \label{fig:workflow}
\end{figure*}

\section{Methodology}
\label{sec:method}

As discussed in Section~\ref{sec:preliminaries}, group-based training methods are efficient because they avoid learning a value model and estimate advantages from rollout groups.
However, they require return variation.
When all rollouts fail, trajectory-level credit becomes uninformative; when all continuations from an anchor state obtain identical returns, anchor-level credit also collapses.
We call these cases \emph{zero-credit groups}.
They are especially common in early long-horizon training, where agents often make partial progress but fail to complete the task.

\methodname{} addresses this failure by learning from \emph{viable failure prefixes}.
As illustrated in Figure~\ref{fig:workflow}, \methodname{} first merges semantically equivalent prefixes into viability regions using Union-Find, then estimates a prefix potential for each region from adaptive milestone progress, empirical future success, and loop statistics.
During optimization, \methodname{} preserves anchor-based comparative credit when it is informative, and routes zero-credit groups to potential-based repair.
The potential signal is strong in the sparse-reward stage and gradually fades as the policy improves.
We analyze why \methodname{} can accelerate early optimization under sparse rewards theoretically in Appendix~\ref{sec:theory}.

\subsection{Viability Potential Estimation}
\label{sec:viability_potential}

We construct a prefix potential that estimates whether an intermediate state remains on a promising path toward task completion.
For a task instruction $x$ and state $s_{i,t}$, we extract a compact progress signature
\begin{equation}
\label{eq:signature}
    z_{i,t}=\psi(x,s_{i,t}),
\end{equation}
where $\psi(\cdot)$ maps raw states to abstract progress descriptors, such as task phase, acquired objects, satisfied constraints, observed evidence, or repeated loop patterns.

\paragraph{Union-Find viability regions.}
Since raw states can be noisy and sparse, \methodname{} groups semantically equivalent prefixes into \emph{viability regions}.
We maintain these regions with Union-Find.
Two state occurrences are merged if they share the same progress signature or satisfy an environment-level equivalence rule:
\begin{equation}
\label{eq:region_equiv}
\begin{aligned}
(i,t)\sim(j,k)
\quad \Longleftrightarrow \quad
 z_{i,t}=z_{j,k} \\
 \text{or } \mathrm{MergeRule}(s_{i,t},s_{j,k})=1 .
\end{aligned}
\end{equation}
The second condition allows the method to merge repeated loop states or states corresponding to the same task milestone beyond exact string equality.
Let $c_{i,t}$ denote the Union-Find representative of occurrence $(i,t)$.
We denote the occurrence set and trajectory set of region $c$ as
\begin{equation}
\label{eq:region_sets}
\begin{aligned}
    \mathcal{O}_k(c)
    &= \{(i,t)\mid c_{i,t}=c\},\\
    \mathcal{T}_k(c)
    &= \{i\mid \exists t,\; c_{i,t}=c\}.
\end{aligned}
\end{equation}

\paragraph{Adaptive milestone potential.}
For each state occurrence, we extract a milestone vector
\begin{equation}
\label{eq:milestone_vector}
    h_{i,t}=(h_{i,t,1},\ldots,h_{i,t,M})\in[0,1]^M,
\end{equation}
where each dimension indicates whether a task-relevant milestone has been reached.
For example, ALFWorld milestones include observing the target object, holding it, completing an operation, or reaching the target receptacle; WebShop milestones include reaching a search page, opening an item page, matching attributes, or purchasing an item.
Details can be found in Appendix~\ref{app:prefix_abstraction}.

Unlike subgoal-reward methods that manually assign dense rewards to predefined milestones, our milestones are not used as fixed rewards.
They are only statistical features for estimating prefix viability, and their importance is learned from rollout data.
Let $Y_{i,t}$ denote a future-progress target, such as future maximum milestone progress or final success.
The marginal utility of milestone $m$ is estimated as
\begin{equation}
\label{eq:milestone_utility}
\begin{aligned}
    \hat{u}_{k,m}
    =
    \Big[
    &\mathbb{E}(Y_{i,t}\mid h_{i,t,m}=1) \\
    &-
    \mathbb{E}(Y_{i,t}\mid h_{i,t,m}=0)
    \Big]_+ .
\end{aligned}
\end{equation}
The normalized milestone weights are updated by EMA:
\begin{equation}
\label{eq:milestone_weight}
    w_{k,m}
    =
    (1-\beta_{\mathrm{amp}})w_{k-1,m}
    +
    \beta_{\mathrm{amp}}
    \frac{\hat{u}_{k,m}}{\sum_{m'}\hat{u}_{k,m'}+\epsilon}.
\end{equation}
Thus, \methodname{} learns which milestones are most predictive of future progress instead of relying on hand-crafted progress rewards.

For each region $c$, we compute its milestone profile, empirical future success, and loop frequency:
\begin{equation}
\label{eq:region_stats}
\begin{aligned}
    \bar{h}_{k,m}(c)
    &=\frac{1}{|\mathcal{O}_k(c)|+\epsilon}
      \sum_{(i,t)\in\mathcal{O}_k(c)}h_{i,t,m}, \\
    \bar{S}_k(c)
    &=\frac{1}{|\mathcal{T}_k(c)|+\epsilon}
      \sum_{i\in\mathcal{T}_k(c)}\mathbb{I}[R_i\ge\epsilon_{\mathrm{succ}}], \\
    \bar{L}_k(c)
    &=\frac{1}{|\mathcal{O}_k(c)|+\epsilon}
      \sum_{(i,t)\in\mathcal{O}_k(c)}\ell_{i,t},
\end{aligned}
\end{equation}
where $\ell_{i,t}$ indicates whether the occurrence belongs to a repeated or looping region.
The viability potential is defined as
\begin{equation}
\label{eq:viability_phi}
\begin{aligned}
    \Phi_k(c)
    =
    \operatorname{Norm}\Big(
    &\sum_{m=1}^{M} w_{k,m}\bar{h}_{k,m}(c) \\
    &+\alpha_s\bar{S}_k(c)-\alpha_l\bar{L}_k(c)
    \Big).
\end{aligned}
\end{equation}
This formulation remains informative even when most rollouts fail, because milestone progress and loop suppression can still distinguish promising prefixes from unproductive ones.

To reduce noise in low-count regions, we smooth the potential toward the batch average:
\begin{equation}
\label{eq:phi_smoothing}
\begin{aligned}
    \tilde{\Phi}_k(c)
    &=\eta_k(c)\Phi_k(c)
    +(1-\eta_k(c))\bar{\Phi}_k,\\
    \eta_k(c)
    &=\frac{|\mathcal{O}_k(c)|}
    {|\mathcal{O}_k(c)|+\lambda_{\mathrm{cnt}}}.
\end{aligned}
\end{equation}
For a transition $(s_{i,t},a_{i,t},s_{i,t+1})$, the potential-based progress advantage is
\begin{equation}
\label{eq:potential_adv}
    A^{\mathrm{pot}}_{i,t}
    =
    \gamma \tilde{\Phi}_k(c_{i,t+1})
    -
    \tilde{\Phi}_k(c_{i,t}).
\end{equation}
It is positive when the transition moves to a more viable region, negative when it regresses or enters a dead end, and close to zero when no meaningful progress is made.
We normalize $A^{\mathrm{pot}}$ within each task group to stabilize scale.

\subsection{Progress-Aware Credit Routing}
\label{sec:credit_routing}

The potential signal should only repair missing comparative credit, not replace reliable anchor-based credit.
For occurrence $(i,t)$, let
\begin{equation}
\label{eq:anchor_stats}
    n_{i,t}=|\mathcal{I}_{s_{i,t}}|,
    \quad
    \sigma_{i,t}
    =
    \operatorname{Std}
    \big(\{G_{j,k}\}_{(j,k)\in\mathcal{I}_{s_{i,t}}}\big).
\end{equation}
We route each occurrence according to the availability of anchor-level return variation:
\begin{equation}
\label{eq:routed_credit}
A^{\mathrm{route}}_{i,t}
=
\begin{cases}
A^{\mathrm{anchor}}_{i,t},
& n_{i,t}>1,\ \sigma_{i,t}>\epsilon_r,\\
\kappa_k \bar{A}^{\mathrm{pot}}_{i,t},
& n_{i,t}>1,\ \sigma_{i,t}\le\epsilon_r,\\
0,
& n_{i,t}=1 .
\end{cases}
\end{equation}
Here, the first case preserves standard anchor credit when repeated anchors contain meaningful return variation; the second case repairs repeated zero-credit anchors with prefix-potential credit; and the third case keeps singleton states neutral because no repeated-state comparison is available.
Thus, potential repair is applied only when comparative anchor evidence exists but collapses due to identical returns.

The coefficient $\kappa_k$ controls the strength of potential repair.
Since potential is most useful before the policy becomes competent, we attenuate it according to relative performance progress.
Let $\bar{s}_k$ be the EMA of batch success rate and $s_0$ be the initial EMA success rate:
\begin{equation}
\label{eq:kappa}
\begin{aligned}
    g_k
    &=
    \operatorname{clip}
    \left(
    \frac{\bar{s}_k-s_0}{1-s_0+\epsilon},0,1
    \right), \\
    \kappa_k
    &=
    \operatorname{clip}(1-g_k,\kappa_{\min},1).
\end{aligned}
\end{equation}
Thus, potential repair is strong when the policy has made little progress and naturally fades as successful rollouts become more frequent.

The final repaired advantage is
\begin{equation}
\label{eq:final_adv}
    \hat{A}_{i,t}
    =
    A_i^{\mathrm{epi}}
    +
    \omega A^{\mathrm{route}}_{i,t},
\end{equation}
where $A_i^{\mathrm{epi}}$ is the episode-level group advantage and $\omega$ controls the contribution of routed step-level credit.

\subsection{Policy Optimization}
\label{sec:method_optimization}

The repaired advantage $\hat{A}_{i,t}$ is assigned to all tokens of textual action $a_{i,t}$.
For token $y_{i,t,\ell}$ with context $c_{i,t,\ell}$, the probability ratio is
\begin{equation}
\label{eq:method_ratio}
    q_{i,t,\ell}(\theta)
    =
    \frac{
    \pi_\theta(y_{i,t,\ell}\mid c_{i,t,\ell})
    }{
    \pi_{\theta_{\mathrm{old}}}(y_{i,t,\ell}\mid c_{i,t,\ell})
    }.
\end{equation}
We optimize the policy with a clipped objective:
\begin{equation}
\label{eq:method_objective}
\begin{aligned}
\mathcal{J}_{\methodname}(\theta)
=
\mathbb{E}_{i,t,\ell}
\Big[
\min\big(
q_{i,t,\ell}(\theta)\hat{A}_{i,t},
\\
\operatorname{clip}(q_{i,t,\ell}(\theta),1-\epsilon_c,1+\epsilon_c)
\hat{A}_{i,t}
\big)
\Big]
\\
-
\beta\,
\mathbb{E}_{i,t,\ell}
\Big[
D_{\mathrm{KL}}
\big(
\pi_\theta(\cdot\mid c_{i,t,\ell})
\Vert
\pi_{\mathrm{ref}}(\cdot\mid c_{i,t,\ell})
\big)
\Big].
\end{aligned}
\end{equation}
The resulting algorithm remains critic-free and does not introduce a learned value model.
Its key difference lies in how the step-level signal is constructed:
\methodname{} preserves comparative credit when return variation is available and repairs zero-credit transitions with a progress-aware potential signal that fades as the policy improves.
\begin{table*}[t]
\centering
\small
\setlength{\tabcolsep}{3.2pt}
\renewcommand{\arraystretch}{1.12}
\caption{
\textbf{Main results on ALFWorld and WebShop.}
For ALFWorld, we report success rates for six subtasks and the overall success rate. For WebShop, we report the average task score and success rate. Results are averaged over 3 random seeds. The best result of each group is in \textbf{bold} and the second-best result is \underline{underlined}. GiGPO$_{\mathrm{w/std}}$ uses $F_{\mathrm{norm}}=\mathrm{std}$, while GiGPO$_{\mathrm{w/o~std}}$ uses $F_{\mathrm{norm}}=1$. \methodname{} uses the $w/std$ setting.
}
\label{tab:main_results}
\resizebox{\textwidth}{!}{
\begin{tabular}{lccccccc|cc}
\toprule
\multirow{2}{*}{\textbf{Method}}
& \multicolumn{7}{c}{\textbf{ALFWorld}} 
& \multicolumn{2}{c}{\textbf{WebShop}} \\
\cmidrule(lr){2-8} \cmidrule(lr){9-10}
& Pick & Clean & Cool & Look & Heat & Pick2 & All & Score & Success \\
\midrule

\rowcolor{gray!15}
\multicolumn{10}{l}{\textit{Closed-source LLMs}} \\
GPT 5.4
& 88.3$_{\pm2.4}$ 
& 52.3$_{\pm8.6}$ 
& 56.9$_{\pm2.0}$ 
& 66.7$_{\pm8.6}$ 
& 70.8$_{\pm20.6}$ 
& 56.3$_{\pm8.4}$ 
& 64.1$_{\pm3.4}$ 
& 9.3$_{\pm1.1}$ 
& 7.0$_{\pm0.6}$ \\
Claude Opus 4.7
& 92.8$_{\pm1.9}$ 
& 81.7$_{\pm3.2}$ 
& 73.6$_{\pm2.0}$ 
& 72.7$_{\pm0.0}$ 
& 72.9$_{\pm10.6}$ 
& 66.2$_{\pm0.6}$ 
& 77.1$_{\pm1.0}$ 
& 23.6$_{\pm6.1}$ 
& 19.8$_{\pm4.8}$ \\
\midrule

\rowcolor{gray!15}
\multicolumn{10}{l}{\textit{Qwen2.5-1.5B-Instruct}} \\
Prompting
& 5.9 & 3.3 & 4.2 & 5.5 & 9.7 & 0.0 & 4.1 & 23.1 & 5.2 \\
ReAct
& 17.4 & 15.7 & 7.7 & 20.5 & 6.2 & 2.0 & 12.8 & 40.1 & 11.3 \\
Reflexion
& 35.3 & 21.7 & 19.4 & 22.2 & 13.6 & 3.7 & 21.8 & 55.8 & 21.9 \\
PPO
& 64.8$_{\pm3.5}$ & 57.1$_{\pm4.9}$ & 46.4$_{\pm4.0}$ & 40.5$_{\pm6.9}$ & 60.6$_{\pm6.6}$ & 47.4$_{\pm1.9}$ & 54.4$_{\pm3.1}$ & 73.8$_{\pm3.0}$ & 51.5$_{\pm2.9}$ \\
RLOO
& 88.3$_{\pm3.0}$ & 71.0$_{\pm5.9}$ & 66.4$_{\pm5.5}$ & 52.8$_{\pm8.6}$ & 62.8$_{\pm8.7}$ & 56.9$_{\pm4.7}$ & 69.7$_{\pm2.5}$ & 73.9$_{\pm5.6}$ & 52.1$_{\pm6.7}$ \\
GRPO
& 85.3$_{\pm1.5}$ & 84.5$_{\pm6.8}$ & 59.7$_{\pm5.0}$ & 53.7$_{\pm8.0}$ & 78.2$_{\pm7.9}$ & 53.5$_{\pm5.6}$ & 72.8$_{\pm3.6}$ & 75.8$_{\pm3.5}$ & 56.8$_{\pm3.8}$ \\
GiGPO$_{\mathrm{w/std}}$
& \underline{94.4$_{\pm5.9}$} 
& \textbf{94.8$_{\pm3.8}$} 
& \underline{79.8$_{\pm4.7}$} 
& 67.5$_{\pm4.6}$ 
& \underline{94.4$_{\pm7.8}$} 
& 76.4$_{\pm5.4}$ 
& \underline{86.7$_{\pm1.7}$} 
& 83.1$_{\pm1.6}$ 
& 65.0$_{\pm3.2}$ \\
GiGPO$_{\mathrm{w/o~std}}$
& \textbf{96.0$_{\pm1.4}$} 
& 91.8$_{\pm5.5}$ 
& 71.7$_{\pm8.4}$ 
& \textbf{76.5$_{\pm3.9}$} 
& 91.3$_{\pm6.3}$ 
& \underline{79.5$_{\pm7.7}$} 
& 86.1$_{\pm4.7}$ 
& \underline{83.5$_{\pm1.8}$} 
& \underline{67.4$_{\pm4.5}$} \\
\rowcolor{gray!10}
\methodname{}
& 92.2$_{\pm1.5}$ 
& \underline{92.4$_{\pm1.5}$} 
& \textbf{89.9$_{\pm2.1}$} 
& \underline{75.6$_{\pm11.0}$} 
& \textbf{94.7$_{\pm0.0}$} 
& \textbf{90.0$_{\pm4.1}$} 
& \textbf{91.1$_{\pm1.0}$} 
& \textbf{84.6$_{\pm1.0}$} 
& \textbf{70.3$_{\pm1.9}$} \\
\midrule

\rowcolor{gray!15}
\multicolumn{10}{l}{\textit{Qwen2.5-7B-Instruct}} \\
Prompting
& 33.4 & 19.3 & 2.8 & 21.6 & 6.9 & 3.2 & 14.8 & 26.4 & 7.8 \\
ReAct
& 48.5 & 34.3 & 18.2 & 35.4 & 13.2 & 17.6 & 31.2 & 46.2 & 19.5 \\
Reflexion
& 62.0 & 44.9 & 36.3 & 41.6 & 30.9 & 23.8 & 42.7 & 58.1 & 28.8 \\
PPO
& 92.3$_{\pm4.0}$ 
& 92.5$_{\pm2.4}$ 
& 80.3$_{\pm2.0}$ 
& 64.0$_{\pm8.4}$ 
& 89.5$_{\pm7.0}$ 
& 68.8$_{\pm8.3}$ 
& 80.4$_{\pm2.7}$ 
& 81.4$_{\pm3.1}$ 
& 68.7$_{\pm5.1}$ \\
RLOO
& 87.6$_{\pm4.3}$ & 87.3$_{\pm5.8}$ & 71.9$_{\pm5.2}$ & 78.2$_{\pm8.3}$ & 81.3$_{\pm7.6}$ & 48.9$_{\pm8.4}$ & 75.5$_{\pm4.6}$ & 80.3$_{\pm3.2}$ & 65.7$_{\pm4.0}$ \\
GRPO
& 90.8$_{\pm5.1}$ & 89.3$_{\pm5.4}$ & 72.5$_{\pm5.4}$ & 66.1$_{\pm6.7}$ & 74.7$_{\pm6.9}$ & 64.7$_{\pm7.3}$ & 77.6$_{\pm5.2}$ & 79.3$_{\pm2.8}$ & 66.1$_{\pm3.7}$ \\
GiGPO$_{\mathrm{w/std}}$
& \underline{97.7$_{\pm1.6}$} 
& \underline{98.8$_{\pm1.6}$} 
& \textbf{89.3$_{\pm8.2}$} 
& 82.7$_{\pm7.9}$ 
& 83.7$_{\pm7.2}$ 
& 79.2$_{\pm6.6}$ 
& \underline{90.8$_{\pm1.3}$} 
& 84.4$_{\pm2.9}$ 
& 72.8$_{\pm3.2}$ \\
GiGPO$_{\mathrm{w/o~std}}$
& 91.8$_{\pm5.4}$ 
& 95.9$_{\pm3.2}$ 
& \underline{86.5$_{\pm5.5}$} 
& \textbf{88.6$_{\pm6.3}$} 
& \underline{90.2$_{\pm2.6}$} 
& \underline{85.2$_{\pm7.5}$} 
& 90.2$_{\pm2.3}$ 
& \underline{86.2$_{\pm2.6}$} 
& \underline{75.2$_{\pm3.8}$} \\
\rowcolor{gray!10}
\methodname{}
& \textbf{100.0$_{\pm0.0}$} 
& \textbf{100.0$_{\pm0.0}$} 
& 68.1$_{\pm2.1}$ 
& \underline{87.8$_{\pm8.8}$} 
& \textbf{91.2$_{\pm2.5}$} 
& \textbf{96.7$_{\pm2.4}$} 
& \textbf{91.9$_{\pm1.3}$} 
& \textbf{86.6$_{\pm2.1}$} 
& \textbf{75.5$_{\pm4.5}$} \\
\bottomrule
\end{tabular}
}
\vspace{-1em}
\end{table*}
\section{Experiments and Results}

In this section, we evaluate whether \methodname{} can improve long-horizon agent training by repairing zero-credit steps with prefix-potential signals. We first report the main results on ALFWorld and WebShop, and then conduct ablations to verify the contribution of each component. Due to the limited space, we evaluate \methodname{} on Search-Augmented QA Tasks in Appendix~\ref{sec:qa_results}.

\subsection{Results on Long-Horizon Agentic Tasks}
\label{sec:main_results}

\noindent \textbf{Experimental setup.}
We evaluate \methodname{} on two long-horizon agentic benchmarks: ALFWorld~\cite{shridhar2021alfworld} and WebShop~\cite{yao2022webshop}.
For ALFWorld, we report the average success rate over six subtasks: Pick, Clean, Cool, Look, Heat, and Pick2.
For WebShop, we report both the average task score and success rate.
All results are averaged over 3 random seeds.
We use Qwen2.5-1.5B-Instruct and Qwen2.5-7B-Instruct~\cite{hui2024qwen2} as trainable base policies, and compare \methodname{} with three groups of baselines: closed-source LLM agents, including \textit{GPT-5.4} and \textit{Claude Opus 4.7}; prompt-based open-source agents, including direct prompting, ReAct~\cite{yao2023react}, and Reflexion~\cite{shinn2023reflexion}; and RL training baselines, including PPO with a critic~\cite{schulman2017proximal}, RLOO~\cite{ahmadian2024back}, GRPO~\cite{shao2024deepseekmath}, and GiGPO~\cite{feng2025group}.
Unless otherwise specified, we follow the training configuration of GiGPO, with a rollout group size of 8 and the same training length for all trainable methods.
For \methodname{}, we use $\omega=0.5$ for routed step-level credit, set $\kappa_{\min}=0.05$ by default and $0.01$ on WebShop, and use $\alpha_{\kappa}=0.05$ for EMA-based performance-progress tracking.
\methodname{} uses the $w/std$ setting for group normalization, following GiGPO$_{\mathrm{w/std}}$.
Full hyperparameter settings and implementation details are provided in Appendix~\ref{app:experimental_details}.
We further analyze training dynamics in Appendix~\ref{sec:training_dynamics} and training-time cost in Section~\ref{sec:training_time}.

\noindent \textbf{Experiment results.}
Table~\ref{tab:main_results} reports the main results on ALFWorld and WebShop.
We summarize three findings.
\textbf{1) \methodname{} accelerates early sparse-reward learning.}
Our method targets the zero-credit regime where GRPO and GiGPO cannot distinguish useful failure prefixes from unproductive ones.
By routing such steps to prefix-potential advantages, \methodname{} provides learning signals before final success becomes frequent.
As shown in Figure~\ref{fig:compare}(c), \methodname{} reaches 40\% ALFWorld success at step 52, earlier than GiGPO at step 62 and GRPO at step 73, showing that viable-prefix learning helps the policy escape the sparse-reward stage faster.
\textbf{2) Better early credit leads to stronger final agents.}
Under the same training length, \methodname{} consistently improves over GiGPO$_{\mathrm{w/std}}$.
With Qwen2.5-1.5B-Instruct, it improves ALFWorld from 86.7\% to \textbf{91.1\%} and WebShop success from 65.0\% to \textbf{70.3\%}.
With Qwen2.5-7B-Instruct, it improves ALFWorld from 90.8\% to \textbf{91.9\%} and WebShop success from 72.8\% to \textbf{75.5\%}.
These gains indicate that the improvement comes from more informative credit construction rather than longer optimization.
\textbf{3) The effect is consistent across environments and scales.}
\methodname{} improves both embodied household control in ALFWorld and web-based decision making in WebShop, and the gains hold for both 1.5B and 7B backbones.
This supports our central hypothesis: long-horizon agent training benefits from identifying viable prefixes inside failed trajectories, rather than treating all zero-return rollouts as equally uninformative.
\subsection{Ablation Study}
\label{sec:ablation}

\noindent \textbf{Experimental setup.}
We conduct ablation studies on WebShop with Qwen2.5-1.5B-Instruct.
All variants follow the same training and evaluation protocol as the main experiments, and results are averaged over 3 random seeds.
We compare four variants: GiGPO$_{\mathrm{w/std}}$, which uses only anchor-based comparative credit; \textbf{w/o GiGPO anchor}, which removes anchor credit and relies only on MVPO prefix-potential repair; \textbf{w/o $\kappa$ decay}, which keeps both anchor and potential branches but disables performance-progress attenuation; and the full \methodname{} with all components enabled.

\begin{table}[t]
\centering
\scriptsize
\setlength{\tabcolsep}{4.2pt}
\renewcommand{\arraystretch}{1.12}
\caption{
\textbf{Ablation study on WebShop with Qwen2.5-1.5B-Instruct.}
We report success rate and task score, averaged over 3 random seeds.
``Anchor'' denotes GiGPO anchor credit, ``Potential'' denotes MVPO prefix-potential credit, and ``$\kappa$ Decay'' denotes performance-progress attenuation.
Best results are \textbf{bolded}.
}
\label{tab:ablation_webshop}
\resizebox{\columnwidth}{!}{
\begin{tabular}{lccccc}
\toprule
\textbf{Method}
& \textbf{Anchor}
& \textbf{Potential}
& \textbf{$\kappa$ Decay}
& \textbf{Success}
& \textbf{Score} \\
\midrule

GiGPO$_{\mathrm{w/std}}$
& \checkmark & -- & --
& \underline{65.0$_{\pm3.2}$}
& 83.1$_{\pm1.6}$ \\

w/o GiGPO anchor
& -- & \checkmark & \checkmark
& 63.5$_{\pm2.1}$
& 79.8$_{\pm0.7}$ \\

w/o $\kappa$ decay
& \checkmark & \checkmark & --
& 60.4$_{\pm4.1}$
& \underline{83.4$_{\pm2.4}$} \\

\rowcolor{gray!10}
\methodname{}
& \checkmark & \checkmark & \checkmark
& \textbf{70.3$_{\pm1.9}$}
& \textbf{84.6$_{\pm1.0}$} \\

\bottomrule
\end{tabular}
}
\vspace{-1.5em}
\end{table}

\noindent \textbf{Experiment results.}
Table~\ref{tab:ablation_webshop} reports the ablation results.
We summarize three findings.
\textbf{1) Potential repair needs anchor routing.}
Removing GiGPO anchor credit reduces success rate from \textbf{70.3\%} to 63.5\% and task score from \textbf{84.6} to 79.8.
This shows that prefix potential is effective for repairing zero-credit steps, but should not replace comparative anchor credit when return variation is available.
Thus, \methodname{} benefits from routing: anchor credit handles reliable comparisons, while potential credit repairs missing signals.
\textbf{2) Potential signals must fade after early learning.}
Removing $\kappa$ decay drops success rate to 60.4\%, even though task score remains 83.4.
This suggests that keeping potential repair fully active throughout training can introduce noisy late-stage guidance.
The result supports our design that prefix potential should mainly help the early sparse-reward stage and gradually give way to anchor-based credit as the policy improves.
\textbf{3) The full method best balances the two signals.}
The full \methodname{} achieves the best success rate and task score, reaching \textbf{70.3\%} and \textbf{84.6}, respectively.
Compared with GiGPO$_{\mathrm{w/std}}$, it improves success by 5.3 points and task score by 1.5 points.
This confirms our central claim: learning from viable failure prefixes is beneficial when potential credit is selectively routed to zero-credit groups and progressively attenuated.

\subsection{Training Time Analysis}
\label{sec:training_time}

\begin{table}[t]
\centering
\scriptsize
\setlength{\tabcolsep}{2.5pt}
\renewcommand{\arraystretch}{1.08}
\caption{
\textbf{Per-step training time.}
\textit{Gen.} denotes rollout generation, \textit{Logp+Ref} denotes old-policy and reference-policy log-probability computation, \textit{Adv.} denotes advantage construction, and \textit{Actor} denotes policy update.
}
\label{tab:training_time}
\resizebox{\columnwidth}{!}{
\begin{tabular}{llccccc}
\toprule
\textbf{Env.} & \textbf{Method} 
& \textbf{Gen.} & \textbf{Logp+Ref} & \textbf{Adv.} & \textbf{Actor} & \textbf{Total} \\
\midrule
\multirow{3}{*}{ALFWorld}
& GRPO
& 183.3 & 19.5 & 0.5 & 35.5 & 283.4 \\
& GiGPO
& 207.9 & 16.6 & 0.7 & 30.7 & 303.9 \\
& \methodname{}
& 227.5 & 20.4 & 1.2 & 35.7 & 340.0 \\
\midrule
\multirow{3}{*}{WebShop}
& GRPO
& 60.0 & 10.7 & 0.1 & 19.6 & 108.1 \\
& GiGPO
& 58.8 & 8.0 & 0.1 & 14.7 & 100.0 \\
& \methodname{}
& 61.2 & 9.4 & 0.3 & 17.3 & 107.9 \\
\bottomrule
\end{tabular}
}
\vspace{-1.5em}
\end{table}

Table~\ref{tab:training_time} reports the per-step wall-clock time under the same Qwen2.5-1.5B training setting. 
For a fair comparison, all methods are profiled under the same environment and hardware setup within each benchmark, i.e., ALFWorld methods are compared under the same ALFWorld setting and WebShop methods under the same WebShop setting. 
We decompose each step into rollout generation (\textit{Gen.}), old-policy and reference-policy log-probability computation (\textit{Logp+Ref}), advantage construction (\textit{Adv.}), and actor update (\textit{Actor}).

The only extra computation introduced by \methodname{} is in the \textit{Adv.} stage, where Union-Find grouping, region statistics, and potential routing are performed. 
Compared with GiGPO, this stage increases by only \(0.5\)s on ALFWorld and \(0.2\)s on WebShop, corresponding to merely \(0.16\%\) and \(0.20\%\) of the total GiGPO step time, respectively. 
Therefore, \methodname{} introduces almost no additional training-time overhead while preserving the efficiency of critic-free group-based RL.

\section{Conclusion}

We identify \emph{zero-credit failure} as a key bottleneck in group-based policy optimization for long-horizon LLM agents: failed rollouts may contain useful prefixes, yet GRPO and GiGPO assign no task-discriminative credit when returns are identical. We propose \methodname{}, which repairs zero-credit steps with prefix-potential advantages estimated over Union-Find viability regions and attenuated by performance progress. Experiments on ALFWorld, WebShop, and search-augmented QA show consistent gains with only \textbf{0.16\%--0.20\%} advantage-computation overhead.

\section*{Limitations}

Although \methodname{} improves sparse-reward agent training by repairing zero-credit steps, it still has two limitations. 
First, \methodname{} relies on meaningful prefix abstraction to estimate viability potentials. 
When observations are highly unstructured or task progress cannot be reliably mapped into region signatures, the Union-Find viability regions may become noisy, reducing the quality of potential-based repair. 
Second, \methodname{} is most beneficial in long-horizon sparse-reward settings where failed trajectories contain recoverable prefixes. 
Its gains may be smaller in short-horizon tasks where final rewards are already frequent or in environments with very weak progress indicators. 
Our experiments focus on ALFWorld, WebShop, and search-augmented QA with Qwen2.5-based agents; broader evaluation on more diverse tool-use, multi-agent, and real-world interactive settings remains future work.

\section*{Ethics Statement}

This work studies reinforcement learning methods for training long-horizon LLM agents in benchmark environments. 
The proposed \methodname{} improves policy optimization by allowing agents to learn from viable prefixes inside failed trajectories, thereby making sparse-reward training more effective. 
Our experiments are conducted on standard research benchmarks, including ALFWorld, WebShop, and search-augmented QA tasks. 
These environments are simulated or benchmarked settings, and our method does not introduce new external tools, collect private user data, or grant agents additional real-world permissions beyond the evaluated tasks.

\noindent \textbf{Dual-use considerations.}
Improving long-horizon agent training can have beneficial applications, such as more reliable web navigation, embodied assistance, tool-use automation, and interactive problem solving. 
However, stronger autonomous agents may also be misused if connected to real-world tools, APIs, websites, or physical systems. 
Potential risks include unsafe tool execution, automated manipulation of online services, unintended goal pursuit, and optimization toward incomplete or misaligned reward signals. 
Because \methodname{} can make agents learn more efficiently from partial progress, it may also improve agents in settings where the intended task is harmful or policy-violating. 
We therefore recommend that deployment of such agents be restricted to well-scoped tasks with explicit safety constraints, access control, sandboxed execution, logging, and human oversight.

\noindent \textbf{Reward and specification risks.}
\methodname{} estimates prefix potentials from rollout statistics and milestone progress. 
Although this improves sparse-reward learning, it may amplify biases or errors in the environment reward, milestone definitions, or progress signatures. 
If the task reward is misspecified, the agent may learn to optimize intermediate states that correlate with success in the benchmark but are unsafe or undesirable in real deployment. 
Thus, prefix-potential credit should be used together with careful reward design, validation on held-out scenarios, and monitoring for reward hacking or shortcut behaviors.

\noindent \textbf{Data and privacy.}
This work does not use private or sensitive user data. 
All evaluations are performed on public or standard research benchmarks. 
For search-augmented QA, the agent interacts with a controlled search interface following benchmark protocols. 
If similar methods are applied to real user data or live web environments, data minimization, privacy protection, consent, and secure logging should be enforced.

\noindent \textbf{Human oversight and deployment.}
Our results should not be interpreted as evidence that trained agents are safe for unrestricted autonomous deployment. 
In real-world applications, agents should operate under explicit permission boundaries, with human approval for high-impact actions and fail-safe mechanisms for abnormal behavior. 
Particularly for domains involving finance, healthcare, legal decisions, cybersecurity, or physical control, additional safety evaluations and domain-specific constraints are necessary before deployment.

\noindent \textbf{Environmental impact.}
Training LLM agents with reinforcement learning can require substantial computation. 
\methodname{} is designed to remain critic-free and introduces only lightweight advantage-construction overhead compared with existing group-based RL methods. 
Nevertheless, large-scale agent training should consider energy efficiency, hardware utilization, and reproducibility, and future work should further study compute-efficient variants.

\bibliography{latex/custom}     

\appendix
\section{Experimental Details}
\label{app:experimental_details}

\subsection{Details of Training}
\label{app:training_details}

\noindent \textbf{Hyperparameters for ALFWorld.}
All methods are configured with the same basic rollout and optimization settings for fair comparison.
The maximum prompt length is 2048 tokens, and the maximum response length is 512 tokens.
Each episode allows up to 50 environment steps.
The actor learning rate is set to $1\times10^{-6}$.
For PPO, which uses an additional critic, the critic learning rate is set to $1\times10^{-5}$.
We adopt the standard sparse task reward: a reward of 10 is assigned for task success and 0 for failure, with a penalty of $-0.1$ for invalid actions.
For group-based RL methods, we use a rollout group size of $N=8$ and sample 16 groups per rollout batch, resulting in $16\times8=128$ parallel environments.
PPO uses 128 separate rollout environments for a comparable batch size.
The rollout temperature is set to 1.0, and the validation temperature is set to 0.4.
The mini-batch size is 256, and the KL-divergence coefficient is 0.01.
For step-level credit methods, the discount factor is set to $\gamma=0.95$.
For \methodname{}, we set the routed step-credit weight to $\omega=0.5$, the minimum potential strength to $\kappa_{\min}=0.05$, and the EMA coefficient for performance-progress tracking to $\alpha_{\kappa}=0.05$.

\noindent \textbf{Hyperparameters for WebShop.}
We use the same general training protocol as ALFWorld.
The maximum prompt length is 4096 tokens, and the maximum response length is 512 tokens.
Each episode is limited to 15 environment steps.
The actor learning rate is $1\times10^{-6}$, and the critic learning rate for PPO is $1\times10^{-5}$.
The task reward is sparse: 10 for success and 0 for failure, with a penalty of $-0.1$ for invalid actions.
All group-based RL methods use a rollout group size of $N=8$ and sample 16 groups per rollout batch, giving 128 parallel environments.
PPO also uses 128 rollout environments.
The rollout temperature is 1.0, and the validation temperature is 0.4.
The mini-batch size is 64, and the KL-divergence coefficient is 0.01.
For step-level credit methods, we set $\gamma=0.95$.
For \methodname{}, we use $\omega=0.5$ and $\alpha_{\kappa}=0.05$.
The minimum potential strength is set to $\kappa_{\min}=0.01$ on WebShop, allowing the potential branch to fade more completely after the policy becomes stronger.

\noindent \textbf{Implementation of \methodname{}.}
For each rollout batch, \methodname{} first extracts progress signatures from environment observations and builds Union-Find viability regions over state prefixes.
For each region, we collect milestone statistics, empirical future-success statistics, and loop statistics to estimate the prefix potential.
The routed advantage is then constructed by preserving anchor-based credit when return variation exists and using potential-based repair when the anchor group is zero-credit.
The performance-progress coefficient $\kappa_t$ is updated from the EMA of batch success rate and controls how strongly the potential branch contributes to the final advantage.
All these computations are performed during advantage construction and do not require an additional value model, reward model, or extra policy forward/backward pass.

\noindent \textbf{Computing details.}
For the reported experiments, Qwen2.5-1.5B models are trained with 8 A100-80G GPUs, and Qwen2.5-7B models are trained with 16 A100-80G GPUs.
Each run is trained for 150 iterations, and all reported numbers are averaged over 3 random seeds.
The 1.5B experiments can also be run with at least 2 A100-80G GPUs by reducing rollout parallelism and using gradient accumulation.
Compared with GRPO and GiGPO, \methodname{} introduces only lightweight CPU-side Union-Find grouping and region-statistics computation during advantage construction.
It does not introduce a critic or additional model inference, so its overall computational cost is nearly identical to existing critic-free group-based RL methods.

\noindent \textbf{Closed-source LLM evaluation.}
We evaluate closed-source LLM agents, including GPT-5.4 and Claude Opus 4.7, as reference models under the same task protocol as open-source agents.
They are not fine-tuned or trained with RL.
For both ALFWorld and WebShop, we use the same evaluation splits, prompt templates, admissible-action interface, and maximum interaction steps as the open-source evaluation.
Specifically, ALFWorld allows at most 50 environment steps and WebShop allows at most 15 steps.
The maximum response length is 512 tokens, and the decoding temperature is set to 0.4, matching the validation setting used for open-source agents.
At each step, the model receives the same task instruction as open-source models.
All closed-source models are evaluated via their official API during May 2026.

\subsection{Prompt Templates}
\label{app:prompt_templates}

\noindent \textbf{ALFWorld prompt.}
For ALFWorld, we use the following prompt template with recent interaction history and admissible actions.

\begin{tcolorbox}[
    enhanced,
    breakable,
    colback=blue!2,
    colframe=blue!45!black,
    coltitle=black,
    colbacktitle=blue!12,
    fonttitle=\bfseries\small,
    title=\textbf{ALFWorld Prompt Template},
    attach boxed title to top left={xshift=2mm,yshift=-2mm},
    boxed title style={
        colback=blue!12,
        colframe=blue!45!black,
        boxrule=0.4pt,
        arc=1.5mm,
        left=1.2mm,
        right=1.2mm,
        top=0.5mm,
        bottom=0.5mm
    },
    boxrule=0.45pt,
    arc=2mm,
    left=1.8mm,
    right=1.8mm,
    top=2.5mm,
    bottom=1.2mm,
    before skip=0.6em,
    after skip=0.8em,
    drop shadow={black!12!white}
]
\small
You are an expert agent operating in the ALFRED Embodied Environment. Your task is to: \{task\_description\}

\vspace{0.5em}
Prior to this step, you have already taken \{step\_count\} step(s). Below are the most recent \{history\_length\} observations and the corresponding actions you took: \{action\_history\}

\vspace{0.5em}
You are now at step \{current\_step\} and your current observation is: \{current\_observation\}

\vspace{0.5em}
Your admissible actions of the current situation are: [\{admissible\_actions\}].

\vspace{0.5em}
Now it's your turn to take an action.

\vspace{0.5em}
You should first reason step-by-step about the current situation. This reasoning process MUST be enclosed within \texttt{<think>} \texttt{</think>} tags.

\vspace{0.5em}
Once you've finished your reasoning, you should choose an admissible action for current step and present it within \texttt{<action>} \texttt{</action>} tags.
\end{tcolorbox}

\noindent \textbf{WebShop prompt.}
For WebShop, we use the following prompt template with recent interaction history and available actions.

\begin{tcolorbox}[
    enhanced,
    breakable,
    colback=teal!2,
    colframe=teal!45!black,
    coltitle=black,
    colbacktitle=teal!12,
    fonttitle=\bfseries\small,
    title=\textbf{WebShop Prompt Template},
    attach boxed title to top left={xshift=2mm,yshift=-2mm},
    boxed title style={
        colback=teal!12,
        colframe=teal!45!black,
        boxrule=0.4pt,
        arc=1.5mm,
        left=1.2mm,
        right=1.2mm,
        top=0.5mm,
        bottom=0.5mm
    },
    boxrule=0.45pt,
    arc=2mm,
    left=1.8mm,
    right=1.8mm,
    top=2.5mm,
    bottom=1.2mm,
    before skip=0.6em,
    after skip=0.8em,
    drop shadow={black!12!white}
]
\small
You are an expert autonomous agent operating in the WebShop e-commerce environment.

\vspace{0.5em}
Your task is to: \{task\_description\}. Prior to this step, you have already taken \{step\_count\} step(s). Below are the most recent \{history\_length\} observations and the corresponding actions you took: \{action\_history\}. You are now at step \{current\_step\} and your current observation is: \{current\_observation\}. Your admissible actions for the current situation are: [\{available\_actions\}].

\vspace{0.5em}
Now it’s your turn to take one action for the current step. You should first reason step-by-step about the current situation, then think carefully which admissible action best advances the shopping goal. This reasoning process MUST be enclosed within \texttt{<think>} \texttt{</think>} tags.

\vspace{0.5em}
Once you’ve finished your reasoning, you should choose an admissible action for current step and present it within \texttt{<action>} \texttt{</action>} tags.
\end{tcolorbox}

\subsection{Implementation Details of Prefix Abstraction}
\label{app:prefix_abstraction}

This section provides the exact prefix abstraction used in our experiments.
Importantly, \methodname{} does not depend on a specific hand-crafted state representation or dense reward design.
The algorithm only requires a lightweight interface that maps raw interaction histories into coarse progress regions and extracts a small set of progress indicators.
In our experiments, we instantiate this interface with deterministic string-matching and structural rules for reproducibility.
These rules can be replaced by other environment metadata, learned state abstractions, or task-specific parsers without changing the policy optimization algorithm.

\noindent \textbf{Progress signatures.}
For each occurrence $(i,t)$, we compute a compact progress signature $z_{i,t}=\psi(x,s_{i,t})$.
The signature is used only to group prefixes into viability regions; it is not used as a reward.

For ALFWorld, we use
\begin{equation}
    \psi(x,s)
    =
    \texttt{loc\_type}
    \;|\;
    \texttt{top2\_objects}
    \;|\;
    \texttt{depth\_bin}.
\end{equation}
Here, \texttt{loc\_type} is extracted from the observation by regular expressions over room or receptacle names, \texttt{top2\_objects} denotes the first two object keywords matched by a fixed vocabulary, and $\texttt{depth\_bin}=3\lfloor t/3\rfloor$.
For example, an observation in the kitchen containing an apple and a knife at depth 6 is mapped to \texttt{kitchen|apple+knife|d6}.

For WebShop, we use
\begin{equation}
    \psi(x,s)
    =
    \texttt{page\_type}
    \;|\;
    \texttt{depth\_bin},
\end{equation}
where \texttt{page\_type} is determined by exact keyword matching:

\begin{center}
\scriptsize
\setlength{\tabcolsep}{3pt}
\renewcommand{\arraystretch}{1.08}
\resizebox{\columnwidth}{!}{
\begin{tabular}{ll}
\toprule
\textbf{Page Type} & \textbf{Extraction Rule} \\
\midrule
\texttt{init} 
& Observation contains \texttt{[search]} \\
\texttt{search\_result} 
& Observation contains \texttt{search results} or \texttt{[back to search]}, without \texttt{[buy now]} \\
\texttt{item\_detail} 
& Observation contains \texttt{[buy now]}, without size/color option buttons \\
\texttt{item\_sub} 
& Observation contains \texttt{[buy now]} and size/color option buttons \\
\texttt{bought} 
& Observation contains \texttt{your order} or \texttt{you have bought} \\
\bottomrule
\end{tabular}
}
\end{center}

For search-augmented QA, the horizon is at most four turns, so we do not apply depth binning.
We use
\begin{equation}
    \psi(x,s)
    =
    \texttt{d}t
    \;|\;
    \texttt{info}_{0/1}
    \;|\;
    \texttt{srch}_{0/1}
    \;|\;
    \texttt{ans}_{0/1}.
\end{equation}
Here, \texttt{info} indicates whether the history contains an \texttt{<information>} block, \texttt{srch} indicates whether the current action contains \texttt{<search>}, and \texttt{ans} indicates whether the current action contains \texttt{<answer>}.
All flags are extracted by exact string matching.

\noindent \textbf{Milestone features.}
For each occurrence, we extract a binary milestone vector $h_{i,t}\in\{0,1\}^{M}$.
These milestones are not optimized as manually assigned dense rewards.
Instead, they serve as progress features whose weights are adapted online by AMP according to their empirical association with future progress.

For ALFWorld, we use five milestones:

\begin{center}
\scriptsize
\setlength{\tabcolsep}{3pt}
\renewcommand{\arraystretch}{1.08}
\resizebox{\columnwidth}{!}{
\begin{tabular}{ll}
\toprule
\textbf{Milestone} & \textbf{Trigger Condition} \\
\midrule
$h_{\mathrm{target}}$ 
& Observation contains \texttt{you pick up} or \texttt{you take} with the target object name \\
$h_{\mathrm{operation}}$ 
& Observation contains \texttt{clean}, \texttt{cool}, \texttt{heat}, \texttt{slice}, or their past-tense forms \\
$h_{\mathrm{place}}$ 
& Observation contains \texttt{you put} or \texttt{you move} \\
$h_{\mathrm{success}}$ 
& Environment reward is positive \\
$h_{\mathrm{invalid}}$ 
& Observation contains \texttt{nothing happens} \\
\bottomrule
\end{tabular}
}
\end{center}

For WebShop, we instantiate the same idea with page-level progress milestones:

\begin{center}
\scriptsize
\setlength{\tabcolsep}{3pt}
\renewcommand{\arraystretch}{1.08}
\resizebox{\columnwidth}{!}{
\begin{tabular}{ll}
\toprule
\textbf{Milestone} & \textbf{Trigger Condition} \\
\midrule
$h_{\mathrm{search}}$ 
& Page type is \texttt{search\_result}, \texttt{item\_detail}, \texttt{item\_sub}, or \texttt{bought} \\
$h_{\mathrm{item}}$ 
& Page type is \texttt{item\_detail}, \texttt{item\_sub}, or \texttt{bought} \\
$h_{\mathrm{option}}$ 
& Page type is \texttt{item\_sub} or \texttt{bought} \\
$h_{\mathrm{success}}$ 
& Environment reward is positive or page type is \texttt{bought} \\
$h_{\mathrm{invalid}}$ 
& Action is invalid or the page does not change after an invalid operation \\
\bottomrule
\end{tabular}
}
\end{center}

For search-augmented QA, we use five tool-use milestones:

\begin{center}
\scriptsize
\setlength{\tabcolsep}{3pt}
\renewcommand{\arraystretch}{1.08}
\resizebox{\columnwidth}{!}{
\begin{tabular}{ll}
\toprule
\textbf{Milestone} & \textbf{Trigger Condition} \\
\midrule
$h_{\mathrm{retrieved}}$ 
& Action contains \texttt{<search>} \\
$h_{\mathrm{relevant}}$ 
& Observation contains a non-empty \texttt{<information>} block longer than 10 characters \\
$h_{\mathrm{new\_info}}$ 
& Same as $h_{\mathrm{relevant}}$ in the current implementation \\
$h_{\mathrm{answer}}$ 
& Action contains \texttt{<answer>} \\
$h_{\mathrm{success}}$ 
& Environment reward is greater than 0.5 \\
\bottomrule
\end{tabular}
}
\end{center}

For ALFWorld and WebShop, the initial milestone weights are
\begin{equation}
    w_0=[0.2,\,0.3,\,0.5,\,10.0,\,0.0].
\end{equation}
These values only initialize the AMP estimator.
During training, AMP updates the weights from rollout statistics, so the effective milestone contribution is learned online rather than fixed by manual reward shaping.

\noindent \textbf{Union-Find merge rule.}
\methodname{} first merges occurrences with identical progress signatures.
In addition, we use a structural cycle rule to merge regions that belong to the same local loop.
For each trajectory, we scan its region signatures in temporal order.
If the same signature is revisited, all regions along the intervening path are merged:
\begin{equation}
\label{eq:merge_rule}
\begin{aligned}
&\text{if } \mathrm{sig}_t=\mathrm{sig}_{t'} \text{ for } t'<t, \\
&\quad \text{then union all adjacent regions on } 
\mathrm{sig}_{t'},\ldots,\mathrm{sig}_{t}.
\end{aligned}
\end{equation}
Equivalently, if an agent leaves a region and later returns to the same signature, the intermediate regions are treated as part of a reversible local loop and share a viability estimate.
This rule is structural and does not use task success labels or manually assigned reward values.

\noindent \textbf{Potential computation.}
For each Union-Find region $C$, we compute
\begin{equation}
\label{eq:app_phi}
    \Phi(C)
    =
    \bar{p}(C)
    +
    \delta
    \big(
    r_{\mathrm{rec}}(C)
    -
    r_{\mathrm{loop}}(C)
    \big),
\end{equation}
where $\bar{p}(C)$ is the average normalized milestone return-to-go over visits to region $C$, $r_{\mathrm{rec}}(C)$ is the fraction of visits after which future milestone progress improves, and
\begin{equation}
    r_{\mathrm{loop}}(C)
    =
    \frac{\texttt{loop\_count}(C)}
    {\texttt{visit\_count}(C)} .
\end{equation}
We set $\delta=0.2$ in all experiments.
The potential is normalized within each rollout batch before computing the potential-difference advantage.

\noindent \textbf{Generality.}
The above rules are used to make the experiments deterministic and reproducible, not to restrict \methodname{} to these environments.
The method itself is agnostic to the specific form of $\psi(\cdot)$ and $h$.
It only assumes that prefixes can be mapped to coarse regions and that some weak progress indicators are available.
Such indicators naturally exist in many agentic settings, including simulator states, admissible actions, browser page types, tool-call traces, retrieved evidence, or learned state embeddings.
Moreover, milestone features are not used as fixed rewards; their weights are adapted online from rollout statistics.
Therefore, \methodname{} should be understood as a general potential-routing framework for repairing zero-credit groups, with the above prefix abstractions serving as one reproducible instantiation.

\section{Additional Results}
\subsection{Training Dynamics Analysis}
\label{sec:training_dynamics}

\begin{figure*}[t]
    \centering
    \begin{subfigure}[t]{0.49\textwidth}
        \centering
        \includegraphics[width=\linewidth]{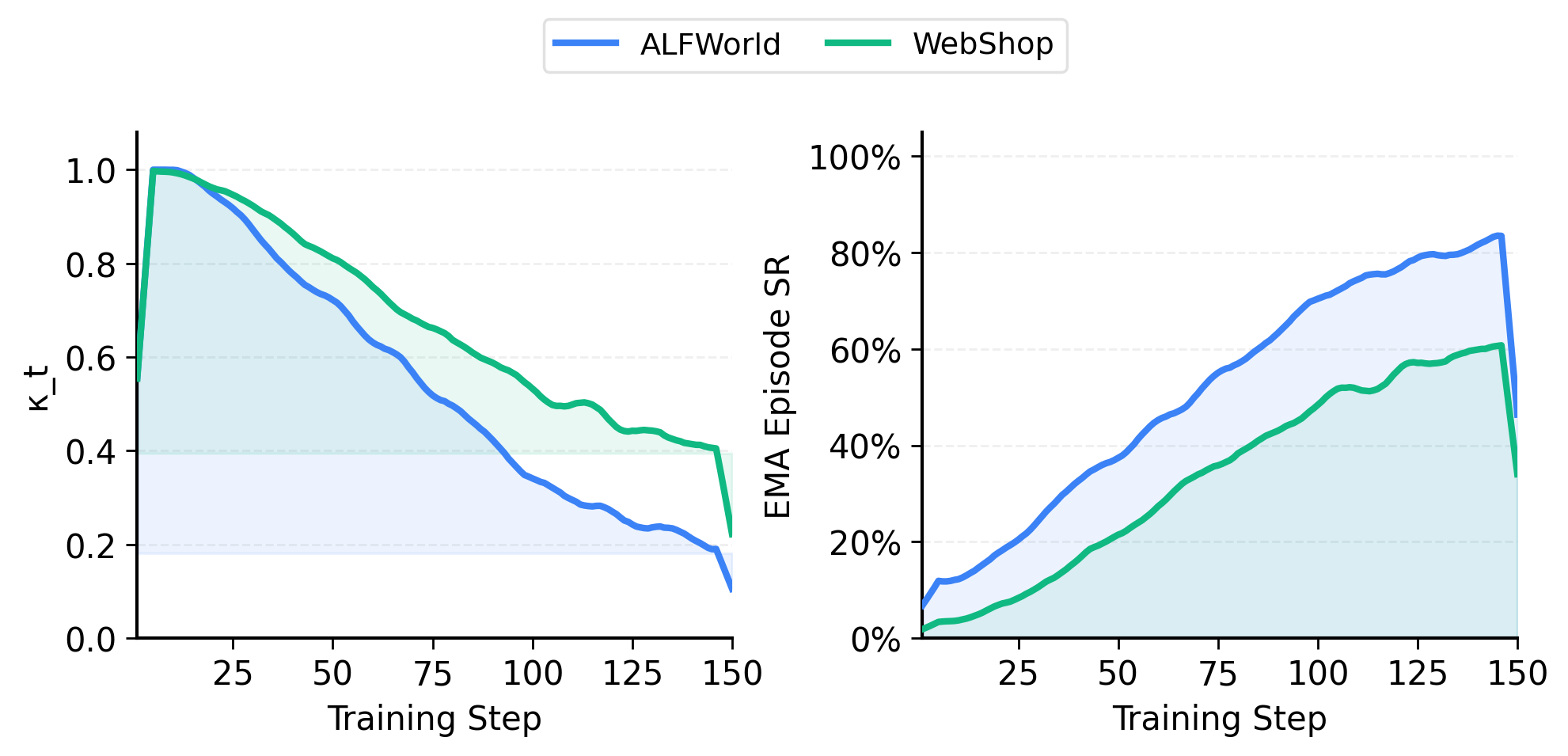}
        \caption{\(\kappa_t\) follows relative performance progress.}
        \label{fig:kappa_decay}
    \end{subfigure}
    \hfill
    \begin{subfigure}[t]{0.49\textwidth}
        \centering
        \includegraphics[width=\linewidth]{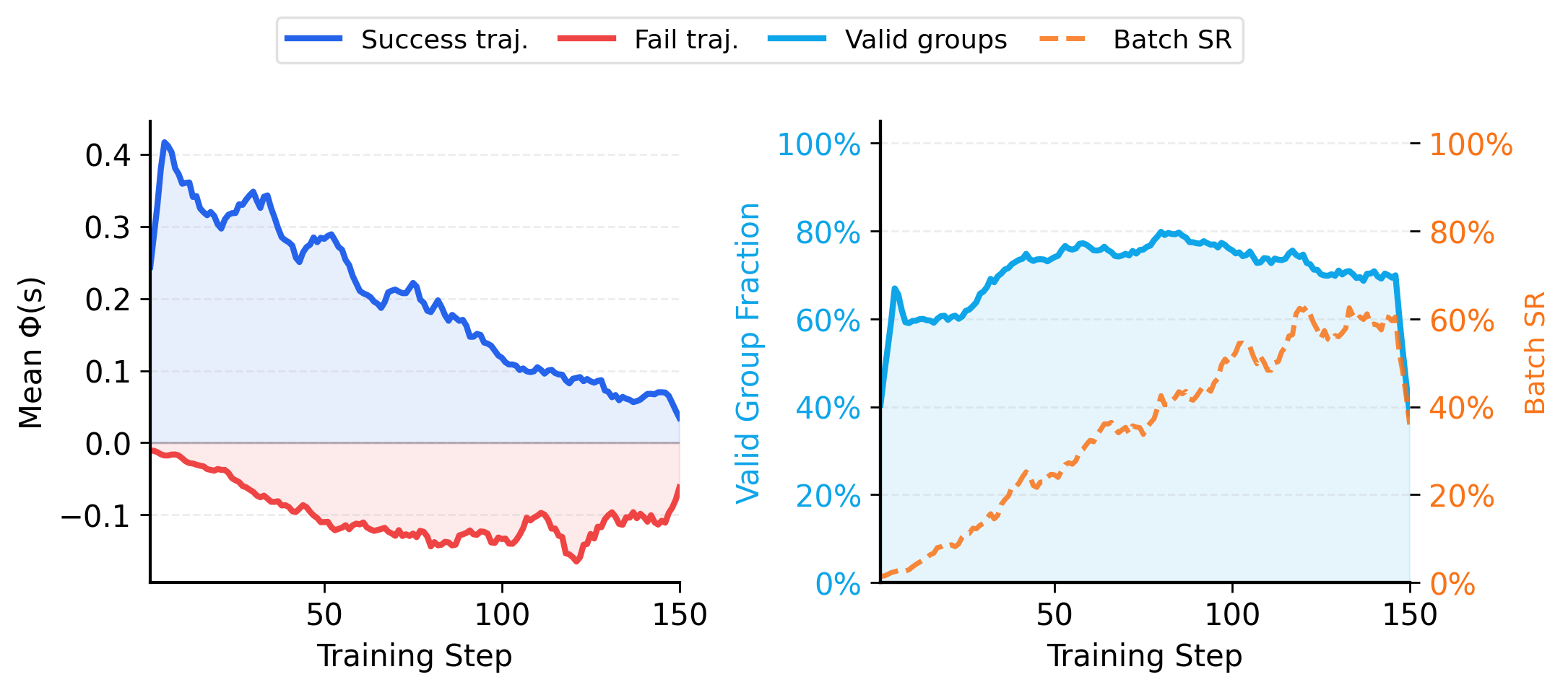}
        \caption{Prefix potential separates successful and failed trajectories.}
        \label{fig:phi_quality}
    \end{subfigure}
    \caption{
    \textbf{Training dynamics of \methodname.}
    \textbf{(a)} The performance-progress attenuation mechanism works as intended. At the beginning of training, the EMA success rate is low and \(\kappa_t\) stays close to 1, allowing prefix-potential credit to repair zero-credit groups with full strength. As the policy improves, the EMA success rate increases and \(\kappa_t\) decreases accordingly, causing the potential branch to fade and leaving more gradient space to anchor-based comparative credit.
    \textbf{(b)} The learned prefix potential remains directionally meaningful throughout training. Successful trajectories maintain higher mean potential than failed trajectories, while the Union-Find region statistics remain stable. This supports the central assumption of \methodname{}: potential can distinguish viable failure prefixes from unproductive prefixes before final rewards become frequent.
    }
    \label{fig:training_dynamics}
    \vspace{-1em}
\end{figure*}

The main motivation of \methodname{} is that early long-horizon agent training contains many zero-credit groups: GRPO and GiGPO cannot distinguish useful failure prefixes from unproductive ones when observed returns have no variation.
\methodname{} repairs this missing signal with prefix-potential credit, but this design requires two properties to hold during training.
First, the potential branch should be strong when the policy is weak, and should gradually fade after the policy becomes competent.
Second, the learned potential should remain directionally correct, assigning higher values to prefixes that are more likely to lead to success.
Figure~\ref{fig:training_dynamics} verifies both properties.

\noindent \textbf{Adaptive retreat of potential credit.}
Figure~\ref{fig:kappa_decay} analyzes the performance-progress attenuation coefficient \(\kappa_t\).
At the beginning of training, the policy is far from competent: the EMA success rate is only about \(13\%\) on ALFWorld and \(3\%\) on WebShop during the first several training steps.
Accordingly, \(\kappa_t\) remains at \(1.0\), meaning that the potential branch is fully active.
This behavior is desirable because zero-credit groups are most frequent in this stage, and the policy needs dense prefix-level signals to learn from partially successful failed trajectories.

As training progresses, the EMA success rate steadily increases.
By step 150, it reaches approximately \(82.5\%\) on ALFWorld and \(61.2\%\) on WebShop.
Following the definition of performance progress, the gain term increases and \(\kappa_t\) decreases to about \(0.20\) on ALFWorld and \(0.40\) on WebShop.
This negative correlation between EMA success rate and \(\kappa_t\) is consistent across both environments, showing that the attenuation mechanism is driven by actual policy improvement rather than by a manually specified training schedule.
In other words, \methodname{} does not require a hand-crafted decay midpoint or fixed step-based schedule: when the policy is weak, potential repair is strong; when the policy becomes stronger, potential repair naturally fades.

This behavior is important for the overall credit-routing design.
Prefix potential is useful for repairing missing credit, but it is not intended to replace anchor-based comparison throughout training.
Once successful continuations become more common, anchor groups contain richer return variation, and GiGPO-style comparative credit becomes more informative.
The decay of \(\kappa_t\) therefore prevents potential signals from dominating late-stage optimization, allowing the final policy update to rely more on direct anchor evidence.
This supports the ablation results in Table~\ref{tab:ablation_webshop}, where disabling \(\kappa\) decay substantially reduces success rate.

\noindent \textbf{Directional quality of prefix potential.}
Figure~\ref{fig:phi_quality} evaluates whether the learned prefix potential \(\Phi(s)\) provides a meaningful progress signal during training.
On WebShop, the mean potential of successful trajectories remains positive throughout training, decreasing from \(+0.44\) at step 1 to \(+0.09\) at step 150.
In contrast, the mean potential of failed trajectories remains negative, from \(-0.02\) at step 1 to \(-0.12\) at step 150.
The sign of this separation never flips, indicating that \(\Phi(s)\) consistently assigns higher viability to prefixes that eventually lead to success and lower viability to prefixes that remain unsuccessful.

This result directly supports the core assumption behind potential-based repair.
When a zero-credit group provides no return variation, \methodname{} uses potential differences to decide whether a transition should receive positive or negative step-level credit.
The observed separation between successful and failed trajectories shows that this signal is directionally aligned with future task completion.
Thus, potential repair is not simply adding dense noise to the advantage estimator; it provides a structured progress signal that distinguishes viable prefixes from dead-end or irrelevant prefixes.

The decreasing absolute gap between successful and failed trajectories is also expected.
The gap shrinks from roughly \(0.46\) early in training to about \(0.20\) near the end.
This does not indicate that the potential becomes invalid.
Instead, as the policy improves, even failed trajectories increasingly visit more plausible states before making their final mistakes.
Consequently, the global potential distribution becomes compressed.
At the same time, \(\kappa_t\) decreases as shown in Figure~\ref{fig:kappa_decay}, reducing the influence of the potential branch precisely when the potential margin becomes smaller.
This coupling between potential quality and potential strength is central to \methodname{}: the method exploits prefix potential when it is most needed and reduces its effect when anchor-based credit becomes more reliable.

\noindent \textbf{Stability of Union-Find viability regions.}
Figure~\ref{fig:phi_quality} also reports the stability of the region construction process.
The batch success rate on WebShop increases from approximately \(2.3\%\) to \(56\%\), while the effective grouping ratio remains stable around \(70\%\)--\(78\%\).
This indicates that the Union-Find viability regions remain usable across the entire training process.
Even when the policy changes and trajectories become more successful, the region abstraction continues to provide sufficient coverage for estimating prefix potential.
This is important because \methodname{} relies on cross-rollout aggregation: individual failed trajectories may be sparse, but semantically equivalent prefixes can be merged into regions with more stable statistics.

Together, these results validate the mechanism of \methodname{}.
Early in training, many rollouts fail and existing group-based methods often produce zero advantages.
During this stage, \(\kappa_t\approx1\) and prefix potential supplies dense repair signals for zero-credit groups.
The potential itself is directionally correct, assigning higher values to prefixes that are more likely to lead to success.
As training proceeds, success becomes more frequent, anchor-based comparative credit becomes more informative, and \(\kappa_t\) automatically decreases.
Therefore, \methodname{} follows the intended behavior: it learns from viable failure prefixes when sparse rewards dominate, and gradually returns control to comparative anchor credit once the policy becomes competent.

\begin{table*}[t]
\centering
\small
\setlength{\tabcolsep}{4.2pt}
\renewcommand{\arraystretch}{1.12}
\caption{
\textbf{Performance on search-augmented QA tasks.}
Models are trained on NQ and HotpotQA with Qwen2.5-3B-Instruct.
$\dagger$ and $*$ indicate in-domain and out-of-domain datasets, respectively.
Best results are in \textbf{bold}.
}
\label{tab:qa_results}
\resizebox{\textwidth}{!}{
\begin{tabular}{lcccccccc}
\toprule
\multirow{2}{*}{\textbf{Method}}
& \multicolumn{3}{c}{\textbf{Single-Hop QA}}
& \multicolumn{4}{c}{\textbf{Multi-Hop QA}}
& \multirow{2}{*}{\textbf{Avg.}} \\
\cmidrule(lr){2-4}
\cmidrule(lr){5-8}
& NQ$^\dagger$ & TriviaQA$^*$ & PopQA$^*$
& HotpotQA$^\dagger$ & 2Wiki$^*$ & MuSiQue$^*$ & Bamboogle$^*$
& \\
\midrule
R1-Instruct
& 27.0 & 53.7 & 19.9
& 23.7 & 29.2 & 7.2 & 29.3
& 27.1 \\
Search-R1
& 34.1 & 54.5 & 37.8
& 32.4 & 31.9 & 10.3 & 26.4
& 32.5 \\
ZeroSearch
& 41.4 & 57.4 & 44.8
& 27.4 & 30.0 & 9.8 & 11.1
& 31.7 \\
StepSearch
& -- & -- & --
& 34.5 & 32.0 & \textbf{17.4} & 34.4
& -- \\
GiGPO
& 42.0 & 59.5 & 42.4
& 36.9 & 37.0 & 12.6 & \textbf{64.1}
& 42.1 \\
\rowcolor{gray!10}
\methodname{}
& \textbf{45.0} & \textbf{61.8} & \textbf{45.7}
& \textbf{37.2} & \textbf{37.1} & 13.4 & 30.4
& \textbf{44.3} \\
\bottomrule
\end{tabular}
}
\end{table*}

\subsection{Evaluation on Search-Augmented QA Tasks}
\label{sec:qa_results}

\noindent \textbf{Experimental setup.}
We further evaluate \methodname{} on search-augmented QA tasks, where an agent must decide when to issue search queries and when to stop with a final answer.
Following prior work, the model is trained on NQ~\cite{kwiatkowski2019natural} and HotpotQA~\cite{yang2018hotpotqa}, and evaluated on both in-domain and out-of-domain datasets.
The single-hop evaluation includes NQ~\cite{kwiatkowski2019natural}, TriviaQA~\cite{joshi2017triviaqa}, and PopQA~\cite{mallen2023not}; the multi-hop evaluation includes HotpotQA~\cite{yang2018hotpotqa}, 2WikiMultiHopQA~\cite{ho2020constructing}, MuSiQue, and Bamboogle~\cite{press2023measuring}.
We use Qwen2.5-3B-Instruct~\cite{hui2024qwen2} as the base policy.
We compare \methodname{} with R1-Instruct~\cite{guo2025deepseek}, Search-R1~\cite{jin2025search}, ZeroSearch~\cite{sun2025zerosearch}, StepSearch~\cite{wang2025stepsearch}, and GiGPO~\cite{feng2025group}.
All results are evaluated at training step 200.
This setting provides a complementary testbed for \methodname{} because search-augmented QA has shorter horizons than ALFWorld and WebShop, but still requires sequential tool-use decisions under sparse final-answer rewards.

\noindent \textbf{Training details.}
The maximum prompt length is 4096 tokens and the maximum response length is 512 tokens.
The maximum interaction turn is 4.
We use a rule-based reward, assigning 1 for a correct final answer and 0 otherwise, with a penalty of $-0.01$ for invalid actions.
The actor learning rate is $1\times10^{-6}$, the training data size is 256, and the rollout group size is 5.
The rollout and validation temperatures are 1.0 and 0.0, respectively.
The mini-batch size is 512, the KL coefficient is 0.001, the step-credit weight is $\omega=1$, and the discount factor is $\gamma=0.95$.
Experiments are trained on 8 A100 GPUs for 200 iterations.

\noindent \textbf{Prompt template.}
We use the following template for search-augmented QA:

\begin{tcolorbox}[
    enhanced,
    breakable,
    colback=pink!3,
    colframe=pink!45!black,
    coltitle=black,
    colbacktitle=pink!16,
    fonttitle=\bfseries\small,
    title=\textbf{Search-Augmented QA Prompt Template},
    attach boxed title to top left={xshift=2mm,yshift=-2mm},
    boxed title style={
        colback=pink!16,
        colframe=pink!45!black,
        boxrule=0.4pt,
        arc=1.5mm,
        left=1.2mm,
        right=1.2mm,
        top=0.5mm,
        bottom=0.5mm
    },
    boxrule=0.45pt,
    arc=2mm,
    left=1.8mm,
    right=1.8mm,
    top=2.5mm,
    bottom=1.2mm,
    before skip=0.6em,
    after skip=0.8em,
    drop shadow={black!10!white}
]
\small
You are an expert agent tasked with answering the given question step-by-step. Your question: \{task\_description\}.

Prior to this step, you have already taken \{step\_count\} step(s). Below is the interaction history where \texttt{<search>} \texttt{</search>} wrapped your past search queries and \texttt{<information>} \texttt{</information>} wrapped the corresponding search results returned by the external search engine. History: \{memory\_context\}

Now it’s your turn to respond for the current step. You should first conduct reasoning process. This process MUST be enclosed within \texttt{<think>} \texttt{</think>} tags. After completing your reasoning, choose only one of the following actions:

(1) If you lack some knowledge, call a search engine using: \texttt{<search>} your query \texttt{</search>}.

(2) If you have enough knowledge to answer confidently, provide your final answer within \texttt{<answer>} \texttt{</answer>} tags, without detailed illustrations. For example, \texttt{<answer>Beijing</answer>}.
\end{tcolorbox}

\noindent \textbf{Experiment results.}
Table~\ref{tab:qa_results} reports the results on search-augmented QA tasks.
Overall, \methodname{} improves the average score from 42.1\% under GiGPO to \textbf{44.3\%}, yielding a +2.2 point gain.
The improvement is especially clear on single-hop out-of-domain datasets: \methodname{} improves TriviaQA from 59.5\% to \textbf{61.8\%} and PopQA from 42.4\% to \textbf{45.7\%}.
It also improves NQ, HotpotQA, 2WikiMultiHopQA, and MuSiQue, indicating that prefix-potential repair can benefit search and answer decisions even in shorter-horizon tool-use tasks.

These results complement the main ALFWorld and WebShop experiments.
Search-augmented QA has fewer interaction turns, so zero-credit prefixes are less severe than in embodied or web-shopping tasks.
Nevertheless, the agent still receives sparse final-answer rewards and must learn which intermediate search actions are useful.
The modest but positive gains suggest that \methodname{} is not limited to environment navigation: its core mechanism, learning from viable prefixes when final outcomes are sparse, also transfers to search-based reasoning agents.
The smaller gain compared with ALFWorld and WebShop is consistent with our story that \methodname{} is most beneficial in longer-horizon settings where failed trajectories contain more recoverable partial progress.

\section{Theoretical Analysis}
\label{sec:theory}

This section explains why \methodname{} can accelerate early optimization under sparse rewards.
The key difference from GRPO and GiGPO is not the policy objective itself, but the construction of the step-level advantage.
When a rollout or anchor group has identical returns, GRPO and GiGPO produce zero task-discriminative credit.
In contrast, \methodname{} repairs such zero-credit steps with a potential-difference signal.
We show that, as long as the prefix potential is positively aligned with future task progress, this repair introduces an additional positive policy-gradient component, yielding a faster expected improvement during the early sparse-reward stage.

\subsection{Zero-Credit Gradient Decomposition}

Let \(J(\theta)\) denote the expected task return of policy \(\pi_\theta\), and let \(g_{\mathrm{B}}\) be the policy-gradient estimator of a baseline group-based method, e.g., GRPO or GiGPO.
For a sampled step \((i,t)\), the baseline advantage is denoted as \(A^{\mathrm{B}}_{i,t}\).
For GRPO, this advantage is shared by all steps in a trajectory; for GiGPO, it further includes anchor-level credit when repeated anchor states have non-zero return variation.

Let \(\mathcal{Z}\) be the set of zero-credit occurrences:
\begin{equation}
\label{eq:zero_credit_set_theory}
    \mathcal{Z}
    =
    \{(i,t)\mid A^{\mathrm{B}}_{i,t}=0
    \text{ due to zero return variation}\}.
\end{equation}
For these steps, the baseline estimator contributes no task-discriminative gradient.
\methodname{} repairs them with a potential advantage
\[
A^{\mathrm{pot}}_{i,t}
=
\gamma\Phi(c_{i,t+1})-\Phi(c_{i,t}),
\]
and forms the routed estimator:
\begin{equation}
\label{eq:mvpo_gradient_decomp}
\begin{aligned}
    g_{\methodname}
    &=
    g_{\mathrm{B}}
    +
    \Delta g_{\mathrm{pot}},\\
    \Delta g_{\mathrm{pot}}
    &=
    \omega\kappa
    \mathbb{E}_{(i,t)\in\mathcal{Z}}
    \big[
    A^{\mathrm{pot}}_{i,t}
    \nabla_\theta \log \pi_\theta(a_{i,t}\mid s_{i,t},x)
    \big].
\end{aligned}
\end{equation}
Thus, \methodname{} differs from GiGPO exactly on the zero-credit region: when anchor credit is informative, it preserves the baseline signal; when the baseline signal collapses, it injects prefix-potential credit.

\subsection{Expected Improvement Bound}

We analyze a single policy update
\[
\theta^+ = \theta + \eta g,
\]
where \(\eta\) is the learning rate.
Assume \(J(\theta)\) is \(L\)-smooth:
\begin{equation}
\label{eq:smoothness}
    J(\theta+\eta g)
    \ge
    J(\theta)
    +
    \eta\langle \nabla J(\theta),g\rangle
    -
    \frac{L\eta^2}{2}\|g\|^2 .
\end{equation}
This standard smoothness inequality says that the improvement is controlled by the alignment between the update direction and the true policy gradient, penalized by a second-order step-size term.

Let
\[
q=\Pr[(i,t)\in\mathcal{Z}]
\]
be the zero-credit step ratio.
We assume that the learned prefix potential is positively aligned with true future progress on zero-credit steps:
\begin{equation}
\label{eq:alignment_assumption}
\begin{aligned}
    \mathbb{E}_{(i,t)\in\mathcal{Z}}
    \Big[
    A^{\mathrm{pot}}_{i,t}
    \langle
    \nabla J(\theta),
    \nabla_\theta \log \pi_\theta(a_{i,t}\mid s_{i,t},x)
    \rangle
    \Big]
    \ge \rho ,
\end{aligned}
\end{equation}
where \(\rho>0\) measures the quality of the potential signal.
This assumption is exactly what Figure~\ref{fig:compare}(b) and Figure~\ref{fig:training_dynamics} support empirically: high-potential prefixes are more likely to lead to future success, and successful trajectories have higher mean potential than failed trajectories.

Let
\[
B_{\mathrm{pot}}
=
\|\Delta g_{\mathrm{pot}}\|,
\quad
B_{\mathrm{B}}
=
\|g_{\mathrm{B}}\|.
\]
Then the one-step improvement of \methodname{} over the baseline satisfies the following bound.

\begin{tcolorbox}[
    colback=gray!4,
    colframe=black!55,
    boxrule=0.5pt,
    arc=1.5mm,
    left=1.5mm,
    right=1.5mm,
    top=1mm,
    bottom=1mm,
    title=\textbf{Proposition 1: Potential Repair Improves Early Update Speed},
    fonttitle=\bfseries\small
]
Under the smoothness condition in Eq.~\eqref{eq:smoothness} and the positive-alignment condition in Eq.~\eqref{eq:alignment_assumption}, the expected one-step improvement of \methodname{} over a baseline group-based estimator satisfies
\begin{equation}
\label{eq:lower_upper_bound}
\begin{aligned}
&\eta\omega\kappa q\rho
-
L\eta^2
\left(
B_{\mathrm{B}}B_{\mathrm{pot}}
+
\frac{1}{2}B_{\mathrm{pot}}^2
\right)
\\
&\le
\mathbb{E}[\Delta J_{\methodname}-\Delta J_{\mathrm{B}}]
\\
&\le
\eta\omega\kappa q\rho_{\max}
+
L\eta^2
\left(
B_{\mathrm{B}}B_{\mathrm{pot}}
+
\frac{1}{2}B_{\mathrm{pot}}^2
\right),
\end{aligned}
\end{equation}
where \(q\) is the zero-credit step ratio, \(\kappa\) is the potential strength, \(\omega\) is the step-credit weight, and \(\rho_{\max}\) upper-bounds the potential-gradient alignment.
\end{tcolorbox}

\noindent \textbf{Interpretation.}
The lower bound shows that the additional gain of \methodname{} increases with three factors:
\textbf{(i)} the zero-credit ratio \(q\), \textbf{(ii)} the potential strength \(\kappa\), and \textbf{(iii)} the alignment quality \(\rho\).
This explains why \methodname{} is most beneficial early in training: zero-credit groups are frequent, \(\kappa\) is close to 1, and potential repair supplies gradients where GRPO and GiGPO provide none.
The second-order term is small when the learning rate is small and the potential gradient is bounded, so the first-order positive term dominates.

The upper bound also clarifies why \methodname{} should not keep potential repair fully active forever.
As training progresses, zero-credit groups become less dominant and anchor-based return variation becomes more informative.
If potential credit remained large, the second-order term could introduce unnecessary late-stage noise.
This motivates the performance-progress attenuation:
\[
\kappa_k=\operatorname{clip}(1-g_k,\kappa_{\min},1),
\]
which reduces \(B_{\mathrm{pot}}\) and the potential contribution as the policy improves.

\subsection{Optimization-Speed Consequence}

Let \(\mu_{\mathrm{B}}\) denote the expected per-step performance increase of a baseline method in a given training phase.
From Proposition~1, the expected per-step increase of \methodname{} is lower-bounded by
\begin{equation}
\label{eq:mvpo_speed}
\begin{aligned}
    \mu_{\methodname}
    \ge
    \mu_{\mathrm{B}}
    +
    \eta\omega\kappa q\rho
    -
    L\eta^2
    \left(
    B_{\mathrm{B}}B_{\mathrm{pot}}
    +
    \frac{1}{2}B_{\mathrm{pot}}^2
    \right).
\end{aligned}
\end{equation}
For any target success level \(S^\star\), let \(S_0\) be the initial success rate.
Ignoring higher-order stochastic fluctuations, the number of training steps required to reach \(S^\star\) is bounded by
\begin{equation}
\label{eq:hitting_time_bound}
\begin{aligned}
    \mu_{\methodname}
    &=
    \mu_{\mathrm{B}}
    +
    \eta\omega\kappa q\rho
    -
    L\eta^2
    \left(
    B_{\mathrm{B}}B_{\mathrm{pot}}
    +
    \frac{1}{2}B_{\mathrm{pot}}^2
    \right), \\
    T_{\methodname}(S^\star)
    &\lesssim
    \frac{S^\star-S_0}{\mu_{\methodname}} .
\end{aligned}
\end{equation}
Compared with the baseline hitting time
\[
T_{\mathrm{B}}(S^\star)\approx \frac{S^\star-S_0}{\mu_{\mathrm{B}}},
\]
\methodname{} reaches the same target faster whenever the positive potential-repair term exceeds the second-order penalty.

\begin{tcolorbox}[
    colback=blue!3,
    colframe=blue!45!black,
    boxrule=0.5pt,
    arc=1.5mm,
    left=1.5mm,
    right=1.5mm,
    top=1mm,
    bottom=1mm,
    title=\textbf{Main Theoretical Takeaway},
    fonttitle=\bfseries\small
]
\methodname{} is expected to update faster than GRPO and GiGPO in the early sparse-reward stage because it converts zero-credit prefixes into aligned potential-gradient signals. The improvement is largest when the zero-credit ratio \(q\) is high and \(\kappa\approx1\), and it naturally fades as \(q\) decreases and performance-progress attenuation reduces \(\kappa\).
\end{tcolorbox}

\subsection{Empirical Verification on ALFWorld}

The theory predicts three observable patterns:
\textbf{(1)} \methodname{} should improve faster in the early stage when zero-credit groups are frequent;
\textbf{(2)} the advantage should be most visible before GiGPO obtains enough successful anchor variation;
\textbf{(3)} late-stage performance should approach GiGPO as \(\kappa\) decays and anchor credit becomes reliable.

Table~\ref{tab:theory_dynamics} verifies these predictions using ALFWorld 1.5B training dynamics.

\begin{table}[t]
\centering
\small
\setlength{\tabcolsep}{4.2pt}
\renewcommand{\arraystretch}{1.12}
\caption{
\textbf{Empirical verification of the theoretical prediction on ALFWorld 1.5B.}
We report validation success rates at representative training steps. \(\Delta_{\mathrm{Gi}}\) denotes the improvement of \methodname{} over GiGPO.
}
\label{tab:theory_dynamics}
\resizebox{\columnwidth}{!}{
\begin{tabular}{c|ccc|c}
\toprule
\textbf{Step} 
& \textbf{GRPO} 
& \textbf{GiGPO} 
& \textbf{\methodname{}} 
& \textbf{$\Delta_{\mathrm{Gi}}$} \\
\midrule
20  & 20.3 & 23.4 & \textbf{28.1} & +4.7 \\
30  & 22.7 & 19.5 & \textbf{32.0} & +12.5 \\
50  & 31.2 & 32.8 & \textbf{47.7} & +14.9 \\
60  & 21.1 & 37.5 & \textbf{57.0} & +19.5 \\
70  & 40.6 & 60.2 & \textbf{60.9} & +0.7 \\
90  & 46.9 & 76.6 & \textbf{78.1} & +1.5 \\
\bottomrule
\end{tabular}
}
\vspace{-0.5em}
\end{table}

The early-stage results strongly match the bound in Eq.~\eqref{eq:lower_upper_bound}.
At step 20, \methodname{} already outperforms GiGPO by \(+4.7\) points.
The gap grows to \(+12.5\) points at step 30, \(+14.9\) points at step 50, and \(+19.5\) points at step 60.
This is exactly the regime where \(q\) is large: many rollouts still fail, making GRPO and GiGPO unable to assign credit to useful prefixes.
\methodname{} instead obtains a positive extra term through potential repair.

As training proceeds, GiGPO begins to receive more informative anchor-level return variation, while the potential branch in \methodname{} is attenuated by performance progress.
Consequently, the gap narrows after step 70.
At step 90, \methodname{} remains slightly ahead of GiGPO (\(78.1\) vs. \(76.6\)).
This behavior is also consistent with the theory: \methodname{} is designed to accelerate the sparse-reward stage rather than permanently override anchor-based credit.
Thus, the empirical dynamics support the theoretical explanation that potential-routed repair improves optimization speed when zero-credit groups dominate, and then fades as comparative anchor credit becomes reliable.

\section{Pseudo Code of \methodname{}}
\label{app:pseudo_code}

Algorithm~\ref{alg:method} presents the training procedure of \methodname{}.

\begin{algorithm*}[t]
    \small
    \caption{\methodname{} Training Procedure}
    \label{alg:method}
    \begin{algorithmic}[1]
        \STATE \textbf{Input:} training task set $\mathcal{D}=\{x\}$,
        policy $\pi_\theta$,
        reference policy $\pi_{\mathrm{ref}}$,
        environment $\mathcal{E}$,
        group size $N$,
        maximum iterations $K$,
        discount factor $\gamma$,
        step-credit weight $\omega$,
        routing threshold $\epsilon_r$,
        AMP update rate $\beta_{\mathrm{amp}}$,
        success and loop weights $\alpha_s,\alpha_l$,
        count-smoothing coefficient $\lambda_{\mathrm{cnt}}$,
        minimum potential strength $\kappa_{\min}$,
        PPO clipping threshold $\epsilon_c$,
        KL coefficient $\beta$.
        
        \STATE \textbf{Initialize:} policy parameters $\theta$,
        milestone weights $\{w_{0,m}\}_{m=1}^{M}$,
        initial success-rate EMA $s_0$,
        current success-rate EMA $\bar{s}_0\leftarrow s_0$.
        
        \STATE \parbox[t]{\linewidth}{\centering \textit{*** \methodname{} training begins ***}}
        
        \FOR{$k=1$ to $K$}
            \STATE Set old policy $\pi_{\theta_{\mathrm{old}}}\leftarrow \pi_\theta$.
            
            \FOR{each task description $x\in\mathcal{D}$}
                \STATE \parbox[t]{\linewidth}{\centering \textit{*** Step A: Group rollout collection. ***}}
                
                \STATE Sample $N$ trajectories $\{\tau_i\}_{i=1}^{N}$ from $\pi_{\theta_{\mathrm{old}}}$ in environment $\mathcal{E}$.
                \STATE Compute trajectory returns $\{R_i\}_{i=1}^{N}$ and batch success rate $\mathrm{SR}_k$.
                \STATE Update success-rate EMA $\bar{s}_k$.
                \STATE Compute episode-level advantages $A_i^{\mathrm{epi}}$ by Eq.~\eqref{eq:grpo_adv}.
                
                \STATE \parbox[t]{\linewidth}{\centering \textit{*** Step B: Anchor construction. ***}}
                
                \FOR{each trajectory $\tau_i$}
                    \FOR{each step $t=1,\ldots,T_i$}
                        \STATE Compute future return $G_{i,t}$ by Eq.~\eqref{eq:future_return}.
                        \STATE Add occurrence $(i,t)$ to anchor set $\mathcal{I}_{s_{i,t}}$.
                    \ENDFOR
                \ENDFOR
                \STATE Compute anchor advantages $A^{\mathrm{anchor}}_{i,t}$ for all anchor groups by Eq.~\eqref{eq:anchor_adv}.
                
                \STATE \parbox[t]{\linewidth}{\centering \textit{*** Step C: Union-Find viability potential estimation. ***}}
                
                \FOR{each trajectory $\tau_i$}
                    \FOR{each step $t=1,\ldots,T_i$}
                        \STATE Extract progress signature $z_{i,t}=\psi(x,s_{i,t})$ by Eq.~\eqref{eq:signature}.
                        \STATE Extract milestone vector $h_{i,t}$ by Eq.~\eqref{eq:milestone_vector}.
                        \STATE Compute loop indicator $\ell_{i,t}$.
                    \ENDFOR
                \ENDFOR
                
                \STATE Initialize Union-Find over all state occurrences.
                \FOR{each pair or candidate pair of occurrences $(i,t),(j,k)$}
                    \IF{$z_{i,t}=z_{j,k}$ or $\mathrm{MergeRule}(s_{i,t},s_{j,k})=1$}
                        \STATE Merge $(i,t)$ and $(j,k)$ in Union-Find by Eq.~\eqref{eq:region_equiv}.
                    \ENDIF
                \ENDFOR
                \STATE Obtain region representative $c_{i,t}$ for each occurrence.
                \STATE Construct $\mathcal{O}_k(c)$ and $\mathcal{T}_k(c)$ for each region by Eq.~\eqref{eq:region_sets}.
                
                \STATE Estimate milestone utilities $\{\hat{u}_{k,m}\}_{m=1}^{M}$ by Eq.~\eqref{eq:milestone_utility}.
                \STATE Update AMP milestone weights $\{w_{k,m}\}_{m=1}^{M}$ by Eq.~\eqref{eq:milestone_weight}.
                
                \FOR{each viability region $c$}
                    \STATE Compute region statistics $\bar{h}_{k,m}(c)$, $\bar{S}_k(c)$, and $\bar{L}_k(c)$ by Eq.~\eqref{eq:region_stats}.
                    \STATE Compute viability potential $\Phi_k(c)$ by Eq.~\eqref{eq:viability_phi}.
                    \STATE Apply count smoothing to obtain $\tilde{\Phi}_k(c)$ by Eq.~\eqref{eq:phi_smoothing}.
                \ENDFOR
                
                \FOR{each occurrence $(i,t)$}
                    \STATE Compute potential advantage $A^{\mathrm{pot}}_{i,t}$ by Eq.~\eqref{eq:potential_adv}.
                \ENDFOR
                \STATE Normalize potential advantages within the task group to obtain $\bar{A}^{\mathrm{pot}}_{i,t}$.
                
                \STATE \parbox[t]{\linewidth}{\centering \textit{*** Step D: Progress-aware credit routing. ***}}
                
                \STATE Compute relative performance gain $g_k$ and potential strength $\kappa_k$ by Eq.~\eqref{eq:kappa}.
                
                \FOR{each occurrence $(i,t)$}
                    \STATE Select the anchor, potential, or neutral branch using $n_{i,t}$ and $\sigma_{i,t}$ in Eq.~\eqref{eq:routed_credit}.
                    \STATE Compute routed credit $A^{\mathrm{route}}_{i,t}$ by Eq.~\eqref{eq:routed_credit}.
                    \STATE Compute repaired advantage $\hat{A}_{i,t}$ by Eq.~\eqref{eq:final_adv}.
                    \STATE Assign $\hat{A}_{i,t}$ to all tokens of action $a_{i,t}$.
                \ENDFOR
                
                \STATE \parbox[t]{\linewidth}{\centering \textit{*** Step E: Policy optimization. ***}}
                
                \STATE Optimize $\pi_\theta$ with the clipped objective in Eq.~\eqref{eq:method_objective}.
                
            \ENDFOR
        \ENDFOR
        
        \STATE \textbf{Output:} trained policy $\pi_\theta$.
    \end{algorithmic}
\end{algorithm*}

\end{document}